\documentclass[10pt]{article}
\usepackage{geometry}
\usepackage{epsfig}
\usepackage[linesnumbered,ruled,vlined]{algorithm2e}
\usepackage{varwidth}
\usepackage{multirow}
\usepackage{graphicx}
\usepackage{epstopdf}
\usepackage{amsmath}
\usepackage{amssymb}
\usepackage{romannum}
\usepackage{amsthm}
\usepackage{booktabs}
\usepackage{float}
\usepackage{subcaption}
\usepackage{caption}
\usepackage{authblk}
\usepackage{abstract}
\usepackage{multicol}
\usepackage{hyperref}
\usepackage{cleveref}
\usepackage[figurename=Fig.]{caption}
\DeclareUnicodeCharacter{2212}{-}
\date{}

\begin{document}
\providecommand{\keywords}[1]{\textit{\textbf{Keywords} - }#1}
\renewcommand{\thepage}{\arabic{page}}
\setcounter{page}{1}
\title{\textbf{Empirical Analysis of Nature-Inspired Algorithms for Autism Spectrum Disorder Detection Using 3D Video Dataset}}
\author[$1, 3$]{Aneesh Panchal}
\author[$2, 3$]{Kainat Khan}
\author[$3, \star$]{Rahul Katarya}

\affil[$1$]{Department of Computational and Data Sciences, Indian Institute of Science, Bangalore, Karnataka$-$560012, INDIA}
\affil[$2$]{Indian Institute of Information Technology, Dharwad, Karnataka$-$580009, INDIA}
\affil[$3$]{Big Data Analytics and Web Intelligence Laboratory, Department of Computer Science $\&$ Engineering, Delhi Technological University, New Delhi$-$110042, INDIA}

\affil[$1$]{Email: aneeshpanchal840850$@$gmail.com}
\affil[$2$]{Email: khankainat388$@$gmail.com}
\affil[$\star$]{Corresponding author:  rahuldtu$@$gmail.com}

\maketitle
\begin{abstract}
Autism Spectrum Disorder (ASD) is a chronic neurodevelopmental condition characterized by repetitive behaviors and impairments in social and communication skills. Despite the clear manifestation of these symptoms, many individuals with ASD remain undiagnosed. This paper proposes a methodology for ASD detection using a three-dimensional walking video dataset, leveraging supervised machine learning classification algorithms combined with nature-inspired optimization algorithms for feature extraction. The approach employs supervised classifiers to identify ASD cases, while nature-inspired optimization techniques select the most relevant features, enhanced by the use of ranking coefficients to identify initial leading particles. This strategy significantly reduces computational time, thereby improving efficiency and accuracy. Experimental evaluation with various algorithmic combinations demonstrates an exceptional classification accuracy of $100\%$ in the best case when using the Random Forest classifier coupled with the Gravitational Search Algorithm for feature selection. The methodology’s application to additional datasets promises improved robustness and generalizability. With its high accuracy and reduced computational requirements, the proposed framework offers significant contributions to both medical and academic fields, providing a foundation for future advances in ASD diagnosis.\\\\
\end{abstract}

\section{Introduction}\label{sec:introduction}
Autism Spectrum Disorder (ASD) is a chronic neurodevelopmental condition characterized by challenges in social communication, restricted interests, and repetitive behaviors. Early diagnosis of ASD is critical, as timely intervention can significantly improve social, emotional, and cognitive outcomes. Conventional diagnostic tools such as the Autism Diagnostic Observation Schedule (ADOS), Childhood Autism Rating Scale (CARS), and Autism Diagnostic Interview-Revised (ADI-R) are widely used in clinical practice. While these instruments are considered reliable, they rely heavily on subjective expert assessment, are time-consuming, and often require specialized clinical training. This has led to delayed or missed diagnoses, particularly in resource-constrained settings.  

To address these limitations, researchers have increasingly turned to computational and data-driven approaches that leverage behavioral, physiological, and multimodal data for ASD detection. In particular, gait and movement patterns are gaining attention as potential digital biomarkers, since motor abnormalities such as atypical stride length, postural instability, or irregular hand-knee coordination are frequently observed in children with ASD. Compared to invasive or clinical assessment procedures, gait-based video analysis is non-intrusive, cost-effective, and scalable, making it well-suited for real-world deployment.

\subsection{Motivation}\label{subsec:motivation}
The prevalence of ASD continues to increase worldwide. Recent estimates suggest that approximately 1 in 68 children in the United States and 1 in 500 children in India are diagnosed with ASD, with rising numbers also being reported across Europe, including France. According to the World Health Organization (WHO), nearly $1$--$1.5\%$ of the global population may fall on the autism spectrum. In India, underdiagnosis remains a critical challenge due to limited clinical infrastructure, whereas in France, recent reports have emphasized the necessity of improved early screening practices within schools and healthcare centers. These circumstances collectively underscore the urgent need for automated, scalable, and interpretable diagnostic support systems.  

Machine Learning (ML) techniques have demonstrated significant potential in detecting ASD across diverse modalities, including eye-gaze patterns, physiological signals, speech, and facial movement data. However, the use of high-dimensional gait video datasets introduces further challenges, such as the curse of dimensionality, feature redundancy, and elevated computational demands. Conventional deep learning architectures—such as Convolutional Neural Networks (CNNs), Recurrent Neural Networks (RNNs), and hybrid CNN-LSTM frameworks—have been applied to pose- and video-based ASD detection tasks. While these approaches have achieved promising results, their effectiveness is often contingent upon access to large-scale training datasets and substantial computational resources. Consequently, their direct applicability to small-scale or resource-constrained clinical datasets, such as the one employed in this study, remains limited.

\subsection{Major contributions}\label{subsec:majorcontri}
The primary objective of this study is to detect ASD in children using supervised ML classification algorithms in combination with nature-inspired optimization techniques. The supervised ML algorithms are employed to perform classification—determining whether a given individual is diagnosed with ASD or not—while the nature-inspired optimization algorithms are utilized for effective feature selection from the dataset. Both supervised learning methods and meta-heuristic optimization algorithms have demonstrated strong capabilities in handling high-dimensional and complex datasets, thereby enhancing predictive performance.  

In particular, the objective of this work can also be framed as the detection of ASD from a three-dimensional walking video dataset of children. Walking represents one of the most fundamental motor skills in human development and carries the potential to reveal early indicators of ASD; thus, gait analysis can serve as a promising modality for diagnostic support. Building on this perspective, the present study leverages features extracted from 3D walking videos of children to address the ASD detection problem.  

In addition to standard ML classification algorithms, we incorporate nature-inspired optimization algorithms for feature selection within high-dimensional datasets. To the best of our knowledge, the work presented here includes several methodological contributions not previously addressed in ASD detection research:  

\begin{itemize}
    \item Integration of nature-inspired optimization algorithms for feature selection and ranking of features using multiple ranking coefficients.
    \item Utilization of three-dimensional walking video data of children as the primary dataset for ASD detection. 
    \item A comprehensive comparative study of various ML classification algorithms, feature-ranking coefficients, and nature-inspired optimization algorithms. 
\end{itemize}

Through these contributions, the proposed methodology seeks to improve both the efficiency and accuracy of current ASD detection models, while simultaneously reducing the total computational cost associated with processing and classification. Thus, this research contributes to the growing body of work on ASD detection by demonstrating the potential synergy between ML-based classifiers and bio-inspired feature selection algorithms. The insights presented here not only establish a foundation for further developments in automated early detection systems but may also support advancements in timely intervention and treatment planning for individuals with ASD.

\subsection{Problem statement}\label{subsec:probstate}
Through a novel methodology that integrates ML classification algorithms with nature-inspired optimization techniques for feature extraction, this study aims to contribute to the field of ASD detection. Nature-inspired optimization algorithms have recently gained considerable attention for feature selection due to their ability to identify the most relevant features while significantly reducing overall computational cost. Furthermore, the incorporation of ranking coefficients within the proposed methodology provides an effective strategy for initializing the leading particles of the optimization algorithm, thereby enhancing efficiency and further reducing computation time.

\subsection{Organization of the paper}\label{subsec:paperorg}
The remaining paper is organized as follows: Section~\ref{sec:litrev} provides a comprehensive literature review of the concepts related to ASD detection using ML algorithms. Section~\ref{sec:prelims} gives the preliminaries required for understanding the subsequent sections of the research paper. Section~\ref{sec:methodology} presents the algorithm and methodology employed in this research paper. Section~\ref{sec:results} describes the dataset specifications, data processing techniques, and feature extraction methods utilizing nature-inspired optimization algorithms. Section~\ref{sec:results} also presents the performance evaluation metrics and results obtained for each algorithm combination. Finally, section~\ref{sec:conclusion} concludes the paper by summarizing the key findings and suggesting potential areas for future research.
\begin{figure}
    \centering
    \resizebox{\columnwidth}{!}{
    \includegraphics{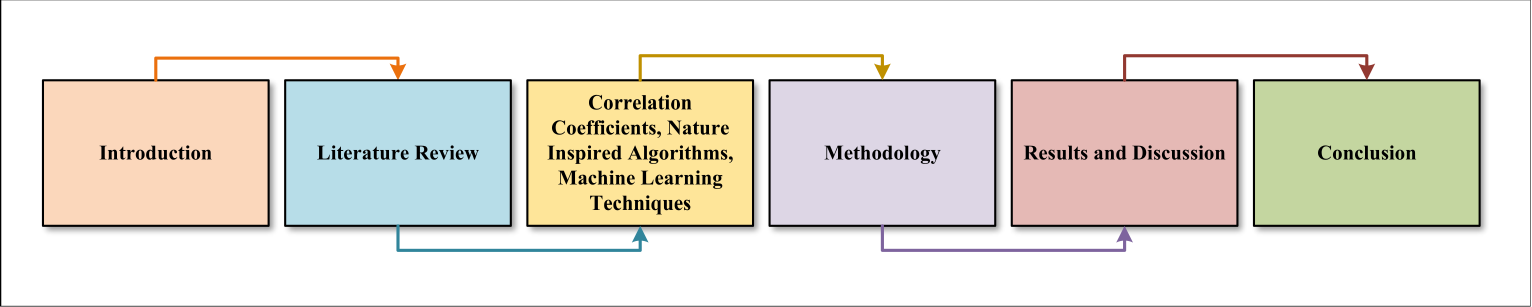}}
    \caption{Organization of the paper}
    \label{fig:paperorg}
\end{figure}

\section{Literature review}\label{sec:litrev}
Early detection of ASD is of paramount importance to ensure timely and appropriate intervention for affected individuals. However, achieving high accuracy in ASD detection remains a significant challenge, requiring advanced assessment tools and specialized expertise. Recent advancements in the field include the following:  

\begin{itemize}
    \item Standardized ASD diagnostic instruments such as the Autism Diagnostic Observation Schedule (ADOS), Childhood Autism Rating Scale (CARS), and Autism Diagnostic Interview-Revised (ADI-R) have been developed to facilitate reliable identification of ASD in individuals~\cite{3,5,6,7}.  
    
    \item Computerized technologies, including robotics, virtual reality, and computer gaming, have shown promising results in enhancing social communication skills among individuals diagnosed with ASD.  
    
    \item Advances in machine learning and mathematical optimization techniques—such as Natural Language Processing (NLP), Long Short-Term Memory (LSTM) networks, autoencoders, and deep learning architectures—have demonstrated superior performance in ASD detection compared to traditional approaches, achieving significantly higher accuracy rates~\cite{3,4,7}.  
    
    \item The utilization of biological and physiological markers, including heart rate data and eye-gaze tracking, has gained attention recently as practical indicators for ASD detection. Notably, systems such as the Autonomous Social Orienting Training System (ASOTS), which employs a Response to Name (RTN) mechanism, are currently in use for diagnostic support~\cite{5,6}.
\end{itemize}

Table~\ref{tab:litrev} summarizes some of the most recent advancements in ASD detection leveraging machine learning techniques. As demonstrated, computerized and algorithmic approaches have become increasingly prominent both in the detection and intervention of ASD. In particular, the effectiveness of nature-inspired optimization algorithms in selecting key features has been shown to reduce overall computational time while enhancing detection accuracy.

\begin{table}
    \tiny
    \centering
    \caption{Literature Review of recently published research papers in related fields.}
    \label{tab:litrev}
    \begin{tabular}{|p{0.5cm}|p{1.25cm}|p{2.25cm}|p{2cm}|p{3cm}|p{2cm}|p{2.25cm}|}\hline 
         \textbf{Year} &  \textbf{Author} &  \textbf{Aim} &  \textbf{Dataset used} &  \textbf{Methodology} &  \textbf{Results} & \textbf{Limitations} \\ \hline 
         2025&  Shin et al.~\cite{shin2025autism}& ASD detection using dual-stream deep learning & Gait 3D video movement dataset~\cite{1} of children.  & Proposed CNN, Transformers, and hybrid CNN+Transformer model  & Accuracy ranging from $93.2\%$ to $95.4\%$; & Computationally expensive and not very reliable due to black-box nature. \\ \hline
         2025&  Zeraati and Davoodi~\cite{zeraati2025LitRev}& ASD-GraphNet: graph learning for ASD diagnosis on fMRI data & Autism Brain Imaging Data Exchange (ABIDE) dataset.  & Graph-based framework including mutual information + PCA and SHAP & Accuracy - $75.25\%$; & External dataset validation and perturbed/noisy data inclusion is required. \\ \hline
         2024&  Colonnese et al.~\cite{colonnese2024bimodal}& Bimodal feature analysis with deep learning for ASD detection & Gait 3D video movement dataset~\cite{1} of children. & Proposed different architectures: LandNet, AngleNet, ConcatNet.  & Accuracy ranging from $97.4\%$ to $98.6\%$; & Computationally expensive and not very reliable due to black-box nature. \\ \hline
         2024&  Kalam et al.~\cite{kalam2024machine}& ML approach for identifying ASD from video & Gait 3D video movement dataset~\cite{1} of children.  & Performed analysis using Random Forest + PCA for feature extraction & Accuracy - $95\%$; & Better method exists for classification accuracy. \\ \hline
         2023&  Henderson et al.~\cite{henderson2023LitRev}& Monocular Human Pose Estimation models & Gait 3D video movement dataset~\cite{1} of children.  & Proposed CLIFF, VideoPose using PoseAug, and PSTMO & Accuracy ranging from $64.96\%$ to $79.87\%$; & Better method exists for classification accuracy. \\ \hline
         
         2022&  Malathi and Kannan~\cite{9}& Adaptive Whale Optimization-based Support Vector Machine for Prediction of ASD  & Child, Adult, and Adolescent ASD Dataset from UCI Machine Learning Website.  & Adaptive Whale Optimization-based SVM is divided into 3 phases: Exploitation, Exploration, and Classification.  & Accuracy for the dataset, Child - $90.753\%$; Adolescent - $90.385\%$; Adult - $89.347\%$; & Better algorithms already exist. Example: Modified Grasshopper Optimization algorithm. \\ \hline 
         2022 &  Sahu and Verma~\cite{11}& Classification of Autistic Spectrum Disorder Using Deep Neural Network with Particle Swarm Optimization  & Child and Adolescent ASD Dataset from UCI Machine Learning Website.  & Particle Swarm Algorithm (PSO) for feature selection and Multi-Layer Perceptron (MLP) and Deep Neural Network (DNN) for classification  & Accuracy for the PSO-DNN model,  Adolescent - $98.077\%$; Child - $100\%$; & Only can be applied to textual data. It is prone to different types of datasets, such as video datasets.  \\ \hline 
         2022 &  Malathi and Kannan~\cite{12}& Robust Artificial Bee Colony Optimization-Based Classifier for Prediction of ASD  & Child, Adult, and Adolescent ASD Dataset from UCI Machine Learning Website.  & Robust Artificial Bee Colony Optimization (RABC) for feature selection and Decision Tree based Gradient Boosting (DTGB) for classification  & Accuracy for the dataset, Child - $82.877\%$; Adolescent - $82.692\%$; Adult - $84.233\%$;  & Better algorithms already exist. Example: Modified Grasshopper Optimization algorithm.  \\ \hline 
         2021 &  Saranya and Anandan~\cite{8}& Autism Spectrum Prognosis using Worm Optimized Extreme Learning Machine (WOEM) Technique  & Kaggle Dataset on Autism Spectrum Disorder. & Glow Worm Algorithm with Extreme Learning Machine (Single Hidden Layer)  & Accuracy - $99\%$; Better results and faster than particle swarm Optimization algorithm  & Only can be applied to textual data. It is prone to different types of datasets, such as video datasets.  \\ \hline 
         2021 &  Elshoky et al.~\cite{13}& Machine Learning Techniques based on Feature Selection for Improving ASD  & Child and Adult ASD Dataset from UCI Machine Learning Website.  & Binary Firefly Feature Selection Algorithm was then applied to various ML algorithms for classification.  & Accuracy - $100\%$; Binary Firefly for feature selection and Logistic Regression.  & can only be applied to textual data. It is prone to different types of datasets, such as video datasets. \\ \hline 
         2021&  Balakrishnan et al.~\cite{14}& Detecting Autism Spectrum Disorder with Sailfish Optimization  & Child and Adult ASD Dataset from UCI Machine Learning Website.  & Random Opposition-based Learning Sailfish Optimization (ROBL-SFO) with 10-fold cross-validation and SVM for classification  & Accuracy for the dataset, Child - $97.30\%$; Adult - $94.20\%$; Efficiency increases by $3.07\%$.  & Not validated for Adolescence dataset. No surety of good results in video/audio dataset. \\ \hline 
         2020& Goel et al.~\cite{15}& Modified Grasshopper Optimization Algorithm for detection of ASD  & Child, Adult and Adolescent ASD Dataset from UCI Machine Learning Website.  & Modified Grasshopper Optimization Algorithm for feature selection. Used 2-3 principal components and trained using Random Forest for classification.  & Accuracy for the dataset, Child - $100\%$; Adolescent - $100\%$; Adult - $99.29\%$;  & Convergence speed is slow for Grasshopper Optimization Algorithm, hence computational time is increased.  \\\hline
         2020& Farrell et al.~\cite{16}& Combined Firefly Algorithm Random Forest (FARF) to Classify Autistic Spectrum Disorders  & Child ASD Dataset from UCI Machine Learning Website.  & Firefly algorithm for feature selection and Random Forest (RF) for classification  & Accuracy - $94.32\%$;  Better than RF with $90.78\%$ accuracy.  & Optimal number of Decision trees can’t be found.  Better algorithm already exists for ASD classification \\\hline
         2020& Li et al.~\cite{7}& Classifying ASD children with LSTM based on raw videos  & Dataset consists of 272 children of primary and special ed. schools (Restricted Data)  & Three three-layer Long Short-Term Memory Network is used for classification.  & Accuracy - $93.4\%$  & Feature extraction may reduce computation time and give satisfactory results. \\\hline
         2019& Abitha and Vennila~\cite{18}& Swarm-based Symmetrical Uncertainty Feature Selection method for ASD  & Child ASD Dataset from UCI Machine Learning Website.  & Combination of Symmetrical Uncertainty and PSO used for feature selection and Naïve Bayes used for classification  & Naïve Bayes, Accuracy - $86.75\%$;  Artificial Neural Networks,  Accuracy - $83.53\%$; & Not validated for Adult and Adolescence dataset. \\\hline
         2018& Samad et al.~\cite{19}& Autism behavioural Markers in Spontaneous Facial, Visual, and Hand Movement Response Data  & Dataset consists of volunteering basis humans (Restricted Data)  & Movement of cursor and Eye gazing on the cursor (Response time analysis) for detection of ASD  & Poor correlation between eye gaze and hand movement in ones who suffers ASD  & Theoretical based paper which can be further improved by using Machine Learning approaches. \\\hline
    \end{tabular}
\end{table}

The primary focus of this research is to address several challenges present in recent advancements in ASD detection, specifically those related to handling high-dimensional datasets, effective feature extraction, and reduction of overall computational time. These challenges with existing methodologies can be summarized as follows:  

\begin{itemize}
    \item The management of high-dimensional datasets remains a significant issue, as many contemporary ASD datasets consist of video and image data characterized by a large number of features. Such high dimensionality increases computational complexity, which can adversely affect both model performance and accuracy. This study aims to develop methods that facilitate efficient analysis and utilization of high-dimensional datasets, ultimately improving ASD detection accuracy.  

    \item Feature extraction constitutes another critical challenge, as the selection of relevant and informative features directly impacts both accuracy and computational efficiency. Effective feature selection algorithms must be capable of identifying the most important features from vast feature spaces, thereby enhancing accuracy while reducing processing time.  

    \item Computational time is a crucial concern, particularly given the urgent need for rapid results in medical applications. This research addresses this challenge by employing appropriate feature selection algorithms and correlation coefficients to identify initial leading particles for optimization-based feature selection, thereby significantly reducing the total computational time.
\end{itemize}

\section{Preliminary}\label{sec:prelims}
This section presents essential background information and fundamental concepts necessary for understanding the subsequent sections of this research paper. It provides detailed descriptions of various correlation coefficients, nature-inspired optimization algorithms, and supervised machine learning classification algorithms employed in this study.

\subsection{Correlation coefficients}\label{subsec:corrcoeff}
In statistics, correlation provides a means to quantify the relationship between two different variables, which may be either random or deterministic. Correlation describes how a change in one variable corresponds to a change in another variable. It can be viewed as a representation of the monotonic association between two variables. In the case of positive correlation, changes in both variables occur in the same direction, whereas in negative correlation, the changes occur in opposite directions~\cite{20}.

\subsubsection{Pearson correlation coefficient}\label{subsubsec:PCC}
The Pearson Correlation Coefficient (PCC), denoted by $\rho$, measures the linear correlation between two variables and ranges between $[-1,1]$, where positive and negative values indicate positive and negative correlations, respectively, and zero indicates no correlation. PCC quantifies both the strength and direction of the linear relationship and can be used for selecting relevant features or important observations from a dataset~\cite{21}.
\par
The PCC between variables $x$ and $y$ is computed as
\begin{equation}\label{eq:PCC}
    \rho_{x,y} = \frac{cov(x,y)}{\sigma_x \sigma_y}
\end{equation}
where $cov(x,y)$ represents the covariance between variables $x$ and $y$, and $\sigma_i$ denotes the standard deviation of the $i^{th}$ variable.

\subsubsection{Spearman correlation coefficient}\label{subsubsec:SCC}
The Spearman Correlation Coefficient (SCC), denoted by $r_S$, also measures the degree of correlation between two variables. Unlike PCC, SCC is less sensitive to outliers and does not assume a linear relationship, making it preferable when dealing with ordinal data or datasets with potential outliers~\cite{21}.
\par
SCC between variables $x$ and $y$ is calculated as 
\begin{equation}\label{eq:SCC}
    r_S = \rho_{R_x,R_y}
\end{equation}
where $R_x$ and $R_y$ denote the ranks of variables $x$ and $y$, respectively.

\subsubsection{Relief ranking coefficient}\label{subsubsec:relief}
The Relief ranking algorithm is an iterative feature weighting method that evaluates the relevance of each feature within a dataset. During each iteration, the algorithm selects instances that are close within the feature space and updates the feature weights accordingly. These weights, known as relief ranking coefficients, help identify the most informative attributes and rank them accordingly. The Relief algorithm is particularly effective for handling high-dimensional datasets~\cite{22}.
\par
The weight vector update for the Relief algorithm is given by
\begin{equation}\label{eq:relief}
    W_i = W_i - (x_i - \text{nearHit}_i)^2 + (x_i - \text{nearMiss}_i)^2
\end{equation}
where $x$ is the feature vector and $W_i$ is the weight for the $i^{th}$ attribute.

\subsection{Nature-inspired algorithms}\label{subsec:natureinspiredalgo}
Nature-inspired algorithms are optimization methods used for a variety of tasks, including optimization, feature selection, and more. Also known as bio-inspired algorithms, they are derived from natural systems and phenomena. These algorithms imitate the dynamic behaviors observed in nature, such as evolutionary processes, swarm intelligence, and genetic inheritance. By leveraging these natural principles, nature-inspired algorithms effectively solve complex and challenging optimization problems~\cite{23,24,25}.

\subsubsection{Gravitational Search Algorithm}\label{subsubsec:GSA}
The Gravitational Search Algorithm (GSA) is a nature-inspired optimization technique based on Newton's law of gravitation. In GSA, each feature is modeled as an object with a corresponding mass, and all objects attract each other via gravitational forces, with heavier masses exerting stronger attraction. This mechanism enables GSA to explore the entire solution space efficiently and converge to an optimal solution. GSA has proven to be a powerful optimization approach for solving complex problems~\cite{26}. Newton's law of gravitation is mathematically expressed as
\begin{equation}\label{eq:GSA}
    F = \sum_{i=1}^{n-1}\sum_{j=i+1}^n G\frac{M_i M_j}{R^2_{i,j}} \qquad \text{and} \qquad a_i = \frac{F}{M_i}
\end{equation}
where $n$ is the total number of objects, $G$ represents the gravitational constant, $F$ denotes the total gravitational force, $R_{i,j}$ is the distance between the $i^{th}$ and $j^{th}$ objects, $M_i$ is the mass of the $i^{th}$ object, and $a_i$ is the acceleration of the $i^{th}$ object.

\subsubsection{Binary Bat Algorithm}\label{subsubsec:BBA}
The Bat Algorithm (BA) is inspired by the echolocation behavior of bats, which use ultrasonic high-frequency sound pulses to navigate and locate prey at night. Bats analyze the reflected echoes to infer information about their surroundings, even in complete darkness. The Binary Bat Algorithm (BBA) is a binary adaptation of the original BA, employing the same principle of echolocation for discrete optimization problems~\cite{27}. The transfer and position update functions in BBA are defined as follows:
\begin{equation}\label{eq:BBA1}
    S(v_i^k(t)) = \frac{1}{1 + e^{-v_i^k(t)}}
\end{equation}
\begin{equation}\label{eq:BBA2}
    x_i^k(t+1) =
    \begin{cases}
    0, & \text{if } \text{rand} < S(v_i^k(t+1)) \\
    1, & \text{if } \text{rand} \geq S(v_i^k(t+1))
    \end{cases}
\end{equation}
where $x_i^k(t)$ and $v_i^k(t)$ represent the position and velocity vectors, respectively, of the $i^{th}$ particle at iteration $t$ in the $k^{th}$ dimension.

\subsubsection{Cuckoo Search Algorithm}\label{subsubsec:CS}
The Cuckoo Search Algorithm (CS) is a nature-inspired optimization algorithm based on the brood parasitic breeding behavior of cuckoo birds, which lay their eggs in the nests of other host birds. CS incorporates Lévy flights, representing random walks with step sizes following a heavy-tailed probability distribution, enabling efficient exploration of the search space and effective escaping from local optima. Consequently, CS demonstrates superior performance compared to many other swarm intelligence algorithms~\cite{28}. The nonlinear update system in CS is expressed as
\begin{equation}\label{eq:CS1}
    x_i^{t+1} = x_i^t + \alpha L(s,\lambda)
\end{equation}
\begin{equation}\label{eq:CS2}
    L(s,\lambda) \sim \frac{\lambda \Gamma(\lambda) \sin\left(\frac{\pi \lambda}{2}\right)}{\pi} \frac{1}{s^{1+\lambda}}, \quad (s > 0)
\end{equation}
where $\alpha > 0$ is a user-defined step size scaling factor.

\subsubsection{Genetic Algorithm}\label{subsubsec:GA}
The Genetic Algorithm (GA) is an evolutionary optimization technique inspired by Darwin's theory of natural selection and the survival of the fittest principle. GA operates iteratively, where the fittest offspring are selected in each generation to produce subsequent generations. Each iteration corresponds to a population generation, consisting of the following stages:

\begin{enumerate}
    \item \textbf{Selection:} Dominant parents are chosen from the population based on fitness, where selection is guided by survival probability and expected fitness, aiming to identify the best features for reproduction.
    
    \item \textbf{Crossover:} This process generates offspring by combining the genetic information of selected parents. The specific crossover methods may vary depending on application needs.
    
    \item \textbf{Repeat the Process:} The selection and crossover steps are repeated across multiple generations to progressively improve the population. If improvement stagnates over generations, mutation is introduced.
    
    \item \textbf{Mutation:} Used sparingly, mutation introduces random changes in offspring to maintain genetic diversity and help avoid local optima when evolutionary progress stagnates.
\end{enumerate}

GA is one of the foundational nature-inspired optimization algorithms. The selection and crossover phases replicate natural selection and genetic recombination, facilitating the survival and propagation of advantageous traits. Through iterative population improvement, GA extensively explores the solution space to identify optimal or near-optimal solutions.

\subsubsection{Grey Wolf Optimization Algorithm}\label{subsubsec:GWO}
Grey Wolf Optimization (GWO) is a nature-inspired algorithm modeled after the hunting behavior of grey wolves, which organize in a hierarchical social structure consisting of four groups: $\alpha$, $\beta$, $\delta$, and $\omega$, where $\alpha$ represents the dominant group and $\omega$ the lowest rank~\cite{29}. While GWO is more complex to implement compared to similar algorithms, it frequently achieves superior results due to its diverse parameter set. The mathematical model of GWO is described as follows:
\begin{equation}\label{eq:GWO1}
    D = |C X_p(t) - X(t)| \omega
\end{equation}
\begin{equation}\label{eq:GWO2}
    X(t+1) = X_p(t) - A D
\end{equation}
\begin{equation}\label{eq:GWO3}
    A = 2 a r_1 - a
\end{equation}
\begin{equation}\label{eq:GWO4}
    C = 2 r_2
\end{equation}
\begin{equation}\label{eq:GWO5}
    a = 2 - \frac{2t}{T_{max}}
\end{equation}
\begin{equation}\label{eq:GWO6}
    \begin{cases}
    D_{\alpha} = |C X_{\alpha} - X| \\
    D_{\beta} = |C X_{\beta} - X| \\
    D_{\delta} = |C X_{\delta} - X|
    \end{cases}
\end{equation}
\begin{equation}\label{eq:GWO7}
    \begin{cases}
    X_1 = X_{\alpha} - A_1 D_{\alpha} \\
    X_2 = X_{\beta} - A_2 D_{\beta} \\
    X_3 = X_{\delta} - A_3 D_{\delta}
    \end{cases}
\end{equation}
\begin{equation}\label{eq:GWO8}
    X(t+1) = \frac{X_1 + X_2 + X_3}{3}
\end{equation}
where $X_p(t)$ and $X(t)$ are the prey and grey wolf position vectors at iteration $t$, $r_1, r_2$ are random vectors in $[0,1]$, $T_{max}$ is the maximum number of iterations, $D_i$ is the distance between the wolf and the $i^{th}$ leading wolf, and $X_i$ for $i \in \{\alpha,\beta,\delta\}$ denote the position vectors of the top three wolves guiding the search.

\subsubsection{Particle Swarm Optimization Algorithm}\label{subsubsec:PSO}
Particle Swarm Optimization (PSO) is a metaheuristic algorithm inspired by the swarming behavior of birds and fish. In PSO, each particle moves through the solution space by updating its velocity based on its own experience and the experience of the entire swarm. The velocity update is influenced by a global best position and the particle’s personal best position, guiding the swarm toward optimal solutions. Iterative updates of velocity and position enable the swarm to converge to the global optimum. The velocity and position of the $i^{th}$ particle are updated as follows:
\begin{equation}\label{eq:PSO1}
    v_i^{t+1} = v_i^t + \alpha \epsilon_1 (g^* - x_i^t) + \beta \epsilon_2 (x_i^* - x_i^t)
\end{equation}
\begin{equation}\label{eq:PSO2}
    x_i^{t+1} = x_i^t + v_i^{t+1}
\end{equation}
where $x_i^t$ and $v_i^t$ are the position and velocity vectors of the $i^{th}$ particle at iteration $t$, $g^*$ is the global best-known position, $x_i^*$ is the personal best position of particle $i$, $\epsilon_1, \epsilon_2$ are independent random vectors uniformly distributed in $[0,1]$, and $\alpha, \beta$ are acceleration coefficients controlling the influence of the social and cognitive components, respectively.

\subsubsection{Whale Optimization Algorithm}\label{subsubsec:WOA}
The Whale Optimization Algorithm (WOA) is inspired by the bubble-net hunting strategy of humpback whales. Whales create spiraling bubbles to concentrate prey into smaller areas before capture. Analogously, WOA progressively narrows the search area around the best solutions, enabling exploitation of the most promising regions of the search space~\cite{30}. The mathematical model of WOA is described by the following equations:
\begin{equation}\label{eq:WOA1}
    D = |C X^*(t) - X(t)|
\end{equation}
\begin{equation}\label{eq:WOA2}
    X(t+1) = X^*(t) - A D
\end{equation}
\begin{equation}\label{eq:WOA3}
    A = 2 a \cdot \text{rand} - a
\end{equation}
\begin{equation}\label{eq:WOA4}
    C = 2 \cdot \text{rand}
\end{equation}
\begin{equation}\label{eq:WOA5}
    a = 2 - \frac{2 t}{T_{max}}
\end{equation}
\begin{equation}\label{eq:WOA6}
    X(t+1) = D' e^{b l} \cos(2 \pi l) + X^*(t)
\end{equation}
\begin{equation}\label{eq:WOA7}
    D' = |X^*(t) - X(t)|
\end{equation}
where $t$ is the current iteration number, $X^*(t)$ is the current best solution (prey position), $X(t)$ is the position vector, $a$ and $b$ are constants, $\text{rand}$ is a random number in $[0,1]$, $T_{max}$ is the maximum number of iterations, $D'$ is the distance between the whale and the prey, $l$ is a random number in $[-1,1]$, and the exponential-cosine term models the logarithmic spiral path of the whale's hunting maneuver. The condition $|A| \geq 1$ governs the exploration phase of the algorithm.

\subsection{Supervised Machine Learning Algorithms}\label{subsec:MLalgo}
Supervised ML algorithms operate on labeled input-output data pairs. These algorithms aim to learn a mapping function or transformation that associates an input variable $x$ with its corresponding output variable $y$. Being data-driven, supervised ML relies entirely on the quality and accuracy of the provided data for effective learning. Once trained, the mapping function enables prediction of output values for unseen inputs, generally achieving higher accuracy than unsupervised learning methods~\cite{31,32}.

\subsubsection{Support Vector Machine}\label{subsubsec:SVM}
Support Vector Machines (SVMs) are supervised ML algorithms applicable to both classification and regression tasks. For classification, SVM constructs an optimal hyperplane that separates data points belonging to different classes, where the hyperplane’s dimension corresponds to the number of features in the dataset.

\subsubsection{K-Nearest Neighbors}\label{subsubsec:KNN}
K-Nearest Neighbors (KNN) is a simple yet effective supervised classification algorithm. It classifies an instance based on the majority class among its $K$ nearest neighbors, determined through Euclidean distance minimization.

\subsubsection{Random Forest}\label{subsubsec:RF}
Random Forest (RF) is a supervised ensemble learning algorithm that builds a large collection of decision trees, either in sequence or in parallel, hence the name. Each tree is trained on a random subset of the data and features, contributing to a robust aggregated prediction.\\

\par
These fundamental supervised ML concepts form the basis for the methodologies proposed in subsequent sections of this paper. The integration of these algorithms with feature selection techniques aims to significantly reduce computational time while improving predictive accuracy.

\section{Methodology}\label{sec:methodology}
To achieve high accuracy in ASD classification, a robust methodological pipeline is designed, consisting of systematic data pre-processing, correlation-based feature ranking, nature-inspired feature selection, model training with nested cross-validation, and comprehensive evaluation. 

\subsection{Data Pre-processing}\label{subsec:preprocess}
The data pre-processing step in the proposed methodology is the first and most important step required to ensure the quality of the dataset. Data pre-processing consists of several minor steps, including data cleaning and null value removal. Data cleaning involved the removal of unnecessary data in the dataset, for example, variance and standard deviation represent the same values. After data cleaning, the dimension of the dataset is reduced to some extent. There are no null values present in our dataset, hence the removal of null value step is not required. Finally, after the data pre-processing step, our dataset was left with $860$ feature columns out of $1259$ feature columns.

\subsection{Main Methodology}

\subsubsection{Feature Ranking}\label{subsec:featureranking}
To prioritize informative biomarkers, three correlation-based criteria rank features, PCC, SCC, and Relief. This multi-perspective ranking highlights the most influential features as leaders for the subsequent nature-inspired optimizers.

\subsubsection{Feature Extraction}\label{subsec:featureextract}
Nature-inspired metaheuristic algorithms are deployed to extract an optimal subset of features. For all optimizers, the initial population is seeded using the top-ranked features, accelerating convergence. The process maintains a balance between classification accuracy ($80\%$ weight) and dimensionality reduction ($20\%$ weight) in the fitness function. We considered seven benchmarked algorithms for this, namely, BBA, CS, GA, GSA, GWO, PSO, and WOA. This step dramatically reduces dataset dimensionality, thereby minimizing computational complexity.

\subsubsection{Classification}\label{subsec:classification}
Selected feature sets are evaluated using three of the most famous machine learning classifiers, KNN with $k=5$, RF with $100$ trees, and SVM with RBF kernel. Classifiers are trained and validated on optimized feature sets for ASD status prediction.

\subsubsection{Nested Cross-Validation, Training and Testing}\label{subsec:trainandtest}
A rigorous nested cross-validation strategy is adopted to prevent overfitting and fine-tune hyperparameters. The outer $5$-fold stratified Cross-Validation (CV) assesses model performance generalization. For each outer fold, feature selection and an inner $3$-fold CV are performed for classifier selection and tuning. Within each fold, the optimizer uses a population of $30$ and a maximum $100$ iterations. Additionally, a standard $80$:$20$ train-test split is maintained, with $20\%$ held out for validation. All hyperparameter searches employ scikit-learn’s \texttt{GridSearchCV} framework.

\subsection{Model Evaluation}\label{subsec:modeleval}
Each combination of ranking method, feature selector, and classifier is evaluated on multiple metrics, including Accuracy, Recall, Precision, and F1-score. Along with these metrics, Confusion matrix analysis, average feature count, and runtime per configuration are also considered for analysis.

Pairwise statistical tests (paired $t$-test, Wilcoxon signed-rank) are also performed across nature-inspired methods to assess the significance of observed improvements. Notably, the combination of Relief, GSA, and RF achieves $100\%$ accuracy with $380$ features, while Relief, CS, and RF provide $96.88\%$ accuracy with only $4$ features, underscoring the strength of dimensionality-driven optimization.

\begin{algorithm}
\caption{Proposed Methodology Pipeline}
\label{algo}
\KwData{Dataset $D \in \mathbb{R}^{m \times n}$ with $m$ samples, $n$ features; target labels $y\in \mathbb{R}^m$}
\KwResult{Performance statistics of classifiers, selected feature subsets, runtime logs, and statistical comparisons}
\vspace{0.5em}
\textbf{Initialize}:\
FeatureMethods $\gets \{\text{PCC},\ \text{SCC},\ \text{Relief}\}$\;
NatureAlgos $\gets \{\text{BBA},\ \text{CS},\ \text{GA},\ \text{GSA},\ \text{GWO},\ \text{PSO},\ \text{WOA}\}$\;
Classifiers $\gets \{\text{KNN},\ \text{RF},\ \text{SVM}\}$\;
\BlankLine
\ForEach{\textnormal{corr} $\in$ FeatureMethods}{
    Compute feature ranking scores $S_{corr}(D, y)$\;
    Initialize the leading particle for NatureAlgo\;
    \ForEach{\textnormal{NatureAlgo} $\in$ NatureAlgos}{
        \ForEach{\textnormal{clf} $\in$ Classifiers}{
            Set up outer stratified $K$-fold CV ($K{=}5$)\;
            $Accuracies \gets [~]$;\ $FeatureCounts \gets [~]$;\ $Times \gets [~]$\;
            \ForEach{(train\_idx, val\_idx) $\in$ outer CV splits}{
                $D_\mathrm{tr}, y_\mathrm{tr} \gets D[\mathrm{train\_idx}], y[\mathrm{train\_idx}]$\;
                $D_\mathrm{val}, y_\mathrm{val} \gets D[\mathrm{val\_idx}], y[\mathrm{val\_idx}]$\;
                Run optimizer $\textnormal{NatureAlgo}$ with population size $30$ and $100$ iterations on $(D_\mathrm{tr}, y_\mathrm{tr})$\;
                Obtain binary feature mask $M \in \{0,1\}^n$\;
                \If{ $\sum M = 0$ }{
                    \textbf{continue} to next fold
                }
                $D_{\mathrm{tr,fs}}, D_{\mathrm{val,fs}} \gets$ select features in $M$\;
                $FeatureCounts \gets FeatureCounts \cup \{\lVert M \rVert_1\}$\;
                Set up stratified inner $L$-fold CV ($L{=}3$)\;
                Perform GridSearchCV on $clf$ hyperparameters with $(D_{\mathrm{tr,fs}}, y_\mathrm{tr})$\;
                Let $clf^*$ = best estimator from tuning\;
                $\hat{y}_\mathrm{val} \gets clf^*(D_{\mathrm{val,fs}})$\;
                $Accuracies \gets Accuracies \cup \{\text{accuracy}(y_\mathrm{val}, \hat{y}_\mathrm{val})\}$\;
                Record runtime for feature selection and inner CV in $Times$\;
            }
            \If{$Accuracies \neq [~]$}{
                Compute $\mu_{\text{acc}}, \sigma^2_{\text{acc}}, 95\%$ CI from $Accuracies$\;
                Compute precision, recall, F1-score, and confusion matrix\;
                Store metrics, mean feature count, and average runtimes into $Results[\textnormal{corr}][\textnormal{NatureAlgo}][\textnormal{clf}]$\;
            }
        }
    }
}
\BlankLine
\textbf{Statistical Analysis}:\;
\ForEach{\textnormal{corr} $\in$ FeatureMethods}{
    \ForEach{\textnormal{clf} $\in$ Classifiers}{
        Compare accuracies of all pairs of NatureAlgos $(w_1, w_2)$ on common folds\;
        Apply paired $t$-test and Wilcoxon signed-rank test between $(Acc_{w_1}, Acc_{w_2})$\;
        Store $p$-values in $Stats[\textnormal{corr}][\textnormal{clf}]$\;
    }
}
\BlankLine
\textbf{Output}: $Results$, $Runtimes$, and $Stats$.
\end{algorithm}

\subsection{Algorithm Analysis}\label{subsec:algoanalysis}
\begin{figure}
     \centering
     \begin{subfigure}[b]{0.3\textwidth}
         \centering
         \includegraphics[width=\textwidth,height=\textwidth]{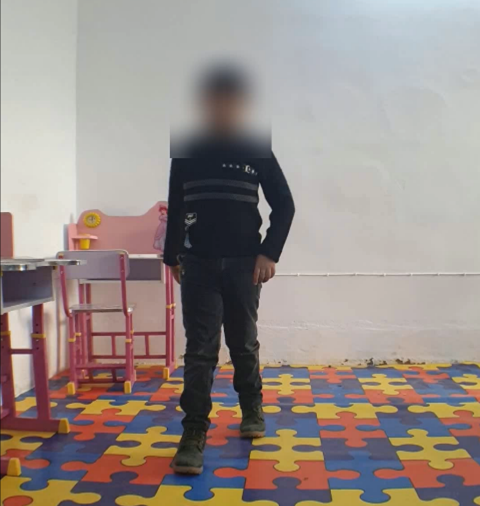}
         \caption{Still from video}
         \label{fig:walking_a}
     \end{subfigure}
     \hfill
     \begin{subfigure}[b]{0.3\textwidth}
         \centering
         \includegraphics[width=\textwidth,height=\textwidth]{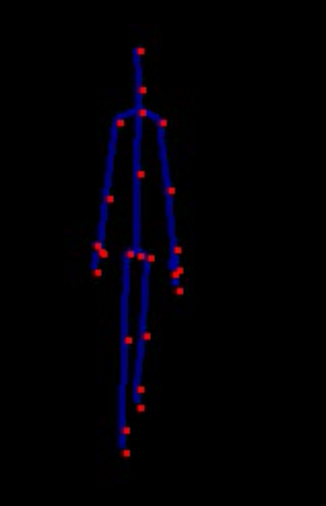}
         \caption{Points taken for study}
         \label{fig:walking_c}
     \end{subfigure}
     \hfill
     \begin{subfigure}[b]{0.3\textwidth}
         \centering
         \includegraphics[width=\textwidth,height=\textwidth]{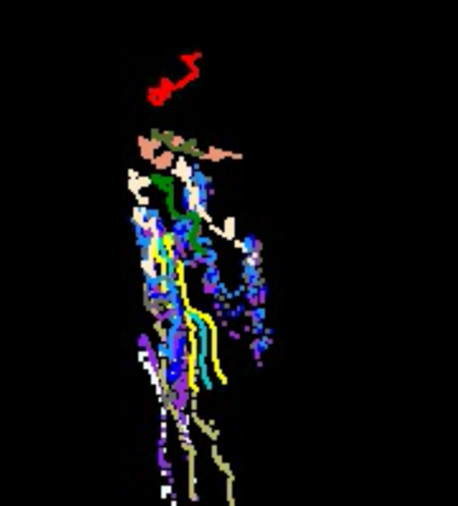}
         \caption{Path followed by points}
         \label{fig:walking_b}
     \end{subfigure}
        \caption{3D Videos of children walking towards the camera}
        \label{fig:walking_images}
\end{figure}

Algorithm~\ref{algo} is designed to handle textual data inputs, which makes it inherently suitable for processing modalities such as video, image, and textual datasets (Section~\ref{subsec:dataset}). This suitability arises from the conversion of raw video or image data into numerical textual representations, enabled by pixel-wise or body landmark analyses. For instance, the original gait video, of which a still frame is shown in Fig.~\ref{fig:walking_a}, captures an individual’s motion. Specific points selected on the body, illustrated in Fig.~\ref{fig:walking_c}, are tracked over time. The trajectories of these points, given by Fig.~\ref{fig:walking_b}, generate the textual dataset used for subsequent analysis and algorithmic processing.

Through feature extraction on this textual data, irrelevant or redundant features are systematically removed to improve computational efficiency and model generalization. The resulting curated dataset forms the basis of the proposed methodology for ASD prediction and optimal model selection.

For example, from Table~\ref{tab:FinalResults}, we can see that the combination of the Relief correlation coefficient, Cuckoo Search (CS) for feature selection, and Random Forest (RF) classifier yields near-optimal performance using only four features with an accuracy of $96.875\%$. These features are,
\begin{enumerate}
    \item \textbf{mean-x-FootRight:} The mean position of the right foot along the x-axis (horizontal).
    \item \textbf{std FoRTThR:} The correlation between the right foot and right thumb.
    \item \textbf{std HaRTKeR:} The correlation between the right hand and right knee.
    \item \textbf{std KeLTSb:} The standard deviation of the left knee.
\end{enumerate}
Three out of these four features explicitly measure deviations or variability from mean body point trajectories, aligning with clinical observations that ASD commonly involves altered motor coordination and neuromotor development. This observation theoretically supports the algorithm’s focus on dynamic and variable features rather than static averages, thereby improving discriminative power.

From an algorithmic perspective, the integration of multi-criteria correlation ranking with nature-inspired feature selection methods enables a balanced trade-off between dimensionality reduction and prediction accuracy. The $80\%$ weighting towards accuracy ensures that selected features robustly characterize ASD-related patterns, while the $20\%$ weighting for feature reduction guards against overfitting and excessive model complexity. Thus, the theoretically informed feature subset and the robust nature-inspired algorithms and classifier algorithm combination identified in this methodology not only facilitate accurate ASD prediction but also promote interpretable models driven by physiologically meaningful markers.

\section{Result analysis and visualization}\label{sec:results}
The result analysis section of this research paper covers various analyses of combinations. Initially, a study of the dataset is done, which provides in-depth knowledge of the data that is used in our methodology. After this, the performance of various combinations of algorithms is evaluated using confusion matrices. Finally, we provide a full explanation of the outcomes we got in the previous steps. Adding up this, these evaluations give us the final performance, statistical analysis, and significance of our methodology required for future research and applications.

\subsection{Dataset description}\label{subsec:dataset}
In order to utilize a video dataset for classification, we must have to transform the dataset into textual data. This is done through analysis of joint motions along all three axes and determining standard deviation as well as variance. Angular deviation of some of the joints and other physical parameters are also extracted as textual data. These extracted features come out to be very useful in determining ASD in an individual, as they determine walking movement patterns, and ASD affects them very severely.
\par
The dataset used in our research is ``\textit{Three-Dimensional Dataset Combining Gait and Full Body Movement of Children with Autism Spectrum Disorders Collected by Kinect v2 Camera Dataset}''~\cite{1}, which is a publicly available dataset. A total of 100 Children are studied in the video dataset, 50 of whom are typical development, and the remaining 50 with ASD detected. A total of 1259 features are initially present in the extracted textual dataset. 
\par
Table~\ref{tab:dataset} shows concise data in the given dataset, which includes the number of children tested for creation of the dataset, the number of data cases present in the given dataset, and the total number of missing values (nil/NA values). Table~\ref{tab:dataset_depth_description} shows all of the features extracted from the 3D walking video dataset, presented with their physical units and the total number of features belonging to that particular set of attributes. Physical units can be used for the data cleaning process in the data pre-processing step in our proposed methodology.
\begin{table}[H]
    \centering
    \caption{General Dataset Description}
    \label{tab:dataset}
    \resizebox{\columnwidth}{!}{
    \begin{tabular}{|c|c|c|c|} \hline 
         \textbf{Data Cases/ Children} &   \textbf{Typical Development} & \textbf{Autism Spectrum Disorder} & \textbf{Total} \\ \hline 
         Number of Children & 50 & 50 & 100 \\ \hline
        Number of Data Cases & 400 & 400 &800 \\\hline
        Missing Values & Nil & Nil &Nil \\\hline
    \end{tabular}}
\end{table}

\begin{table}
    \centering
    \caption{Features included in dataset (.xlsx extracted from .avi)~\cite{1}}
    \label{tab:dataset_depth_description}
    \resizebox{\columnwidth}{!}{
    \begin{tabular}{|c|c|c|c|c|} \hline 
         \textbf{Abbreviation} &  \textbf{Description} &  \textbf{Type} &  \textbf{Unit} & \textbf{Features} \\ \hline 
         mean-coordination-Joint &  Mean of x, y, and z coordination of 25 body joints &  Floating-point & m & 75 \\ \hline 
         variance-coordination-Joint &  Variance of x, y, and z coordination of 25 body joints &  Floating-point &  m2 & 75 \\ \hline 
         std-coordination-Joint  &  Standard deviation of x, y, and z coordination of 25 body joints &  Floating-point &  m & 75 \\ \hline 
         mean Angle &  Mean of 18 angles between body joints &  Floating-point   &  deg & 18 \\ \hline 
         variance Angle  &  Variance of 18 angles between body joints &  Floating-point   &  deg2 & 18 \\ \hline 
         std Angle  &  Standard deviation of 18 angles between body joints &  Floating-point   &  deg & 18 \\ \hline 
         mean joint1 T Joint2  &  Mean of distance between body joints &  Floating-point   &  m & 24 \\ \hline 
         variance joint1 T Joint2  &  Variance of distance between body joints &  Floating-point   &  m2 & 24 \\ \hline 
         std joint1 T Joint2  &  Standard deviation of distance between body joints &  Floating-point   &  m & 24 \\ \hline 
         Mean Joint TGr  & Mean of distance between Joints and ground & Floating-point   & m &282 \\ \hline 
         variance Joint TGr  & Variance of distance between Joints and ground & Floating-point   & m2 &282 \\ \hline 
         std Joint1 TGr  & Standard deviation of distance between Joints and ground & Floating-point   & m &282 \\ \hline 
         Rom Joint coordination  & Range of movement of 25 joint on x and y coordination & Floating-point   & m &50 \\ \hline 
         MaxStLe & Maximum stride length & Floating-point   & m &1 \\ \hline 
         MaxStWi & Maximum stride width & Floating-point   & m &1 \\ \hline 
         StrLe & Stride length & Floating-point   & m &1 \\ \hline 
         GaCT & Gait cycle time & Time   & ms &1 \\ \hline 
         StaT & Stance time & Time   & ms &1 \\ \hline 
         SwiT & Swing time  & Time   & ms &1 \\ \hline 
         Velocity & Velocity & Floating-point   & m/s &1 \\ \hline 
         HaTiLPos & Hand Tip left position relative to SpanBase & Binary (0, 1) & - &1 \\ \hline 
         HaTiRPos & Hand Tip Right position relative to SpanBase & Binary (0, 1) & - &1 \\ \hline 
         MaxDBFE and MinDBFE & Maximum and minimum distance between feet & Floating-point & m &2 \\ \hline 
         Threshold & Using in extract one gait cycle = average of distance between feet & Floating-point & m &1 \\ \hline 
         Total Number of Features & - & - & - &1259 \\ \hline
    \end{tabular}}
\end{table}

\subsection{Performance/ Evaluation metrics}\label{subsec:metrics}
In this research work, we evaluated the performance of the model (combination of algorithms) using confusion matrices. A confusion matrix is used to describe performance, which encompasses various variables, including True Positive, False Positive, True Negative, and False Negative~\cite{13}. These matrices give in-depth knowledge of the model. Hence, the evaluation of the classification model is done using test accuracy, recall, precision, and F1 score.

\subsection{Results}\label{subsec:results}
All the experiments are performed on an Intel $i9$-$14900$K CPU with a $3.20$GHz frequency, $32$GB RAM, and $32$ cores. Base codes for implementation (Base implementation of nature-inspired algorithms for feature selection and Relief ranking coefficient) are adapted from the Py\_FS Python package~\cite{44} (Base implementation of nature-inspired algorithms for feature selection and Relief ranking coefficient) and implemented with Python $3.10.13$. All parameters for the nature-inspired algorithms were set to their default values as specified in the Py\_FS Python package.

After comprehensive evaluation of various combinations involving correlation coefficients, nature-inspired optimization algorithms for feature selection, and supervised machine learning classifiers (provided in \textbf{Appendix}), we identified that the combination of the GSA and RF yielded the highest classification accuracy of $100\%$. This optimal configuration employed the Relief ranking coefficient and selected $380$ features. Significantly, among all tested combinations, this (Relief, GSA, RF) setup uniquely achieved perfect accuracy while simultaneously reducing features by $69.817\%$ in the best case. If the priority shifts towards maximal feature reduction without placing paramount emphasis on attaining perfect accuracy, the combination of Relief coefficient, CS optimizer, and Random Forest classifier offers a notable alternative. This setup achieved an accuracy of $96.875\%$ using only $4$ features out of a total of $1259$, corresponding to an exceptional feature reduction of $99.68\%$ in the best case. For scenarios where computational efficiency and timing are critical, the combination of SCC, GA, and the KNN classifier stands out. This trio delivered an accuracy of $98.75\%$ using $399$ features with a runtime of merely $5$ seconds. Remarkably, this runtime is reduced by $87.18\%$ compared to the (Relief, GSA, RF) combination and by $85.2\%$ relative to the (Relief, CS, RF) setup. Hence, depending on specific requirements - whether prioritizing accuracy, feature reduction, or computational efficiency - these three options provide flexible choices. Table~\ref{tab:comprehensive_results} summarizes these key results for ease of comparison.

\begin{figure}
    \centering
    \includegraphics[width=0.75\textwidth]{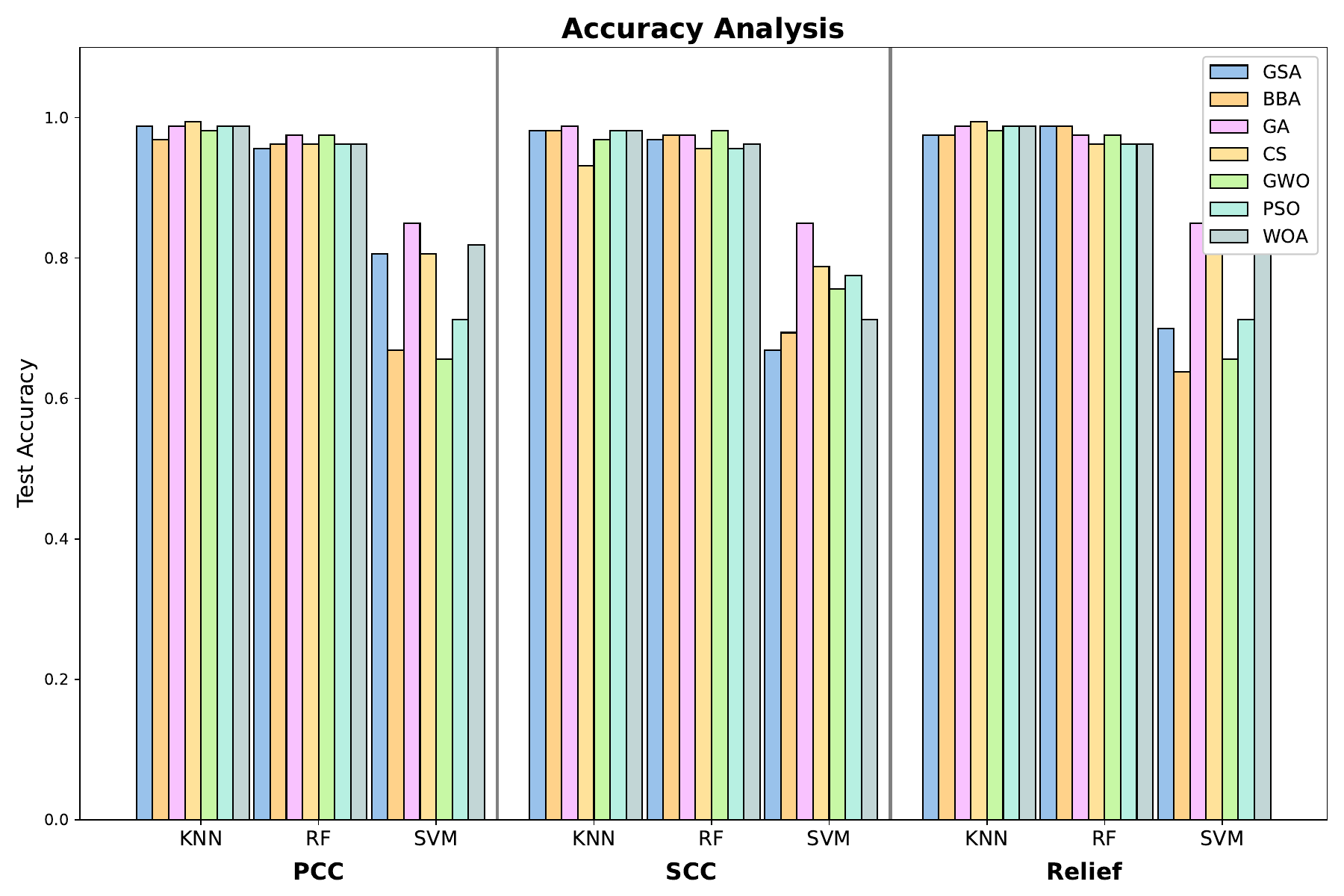}
    \caption{Avg. test accuracy analysis of various combinations of feature extraction and classification algorithms}
    \label{fig:accuracy}
\end{figure}

\begin{figure}
    \centering
     \begin{subfigure}[b]{0.32\textwidth}
         \centering
         \includegraphics[width=\textwidth]{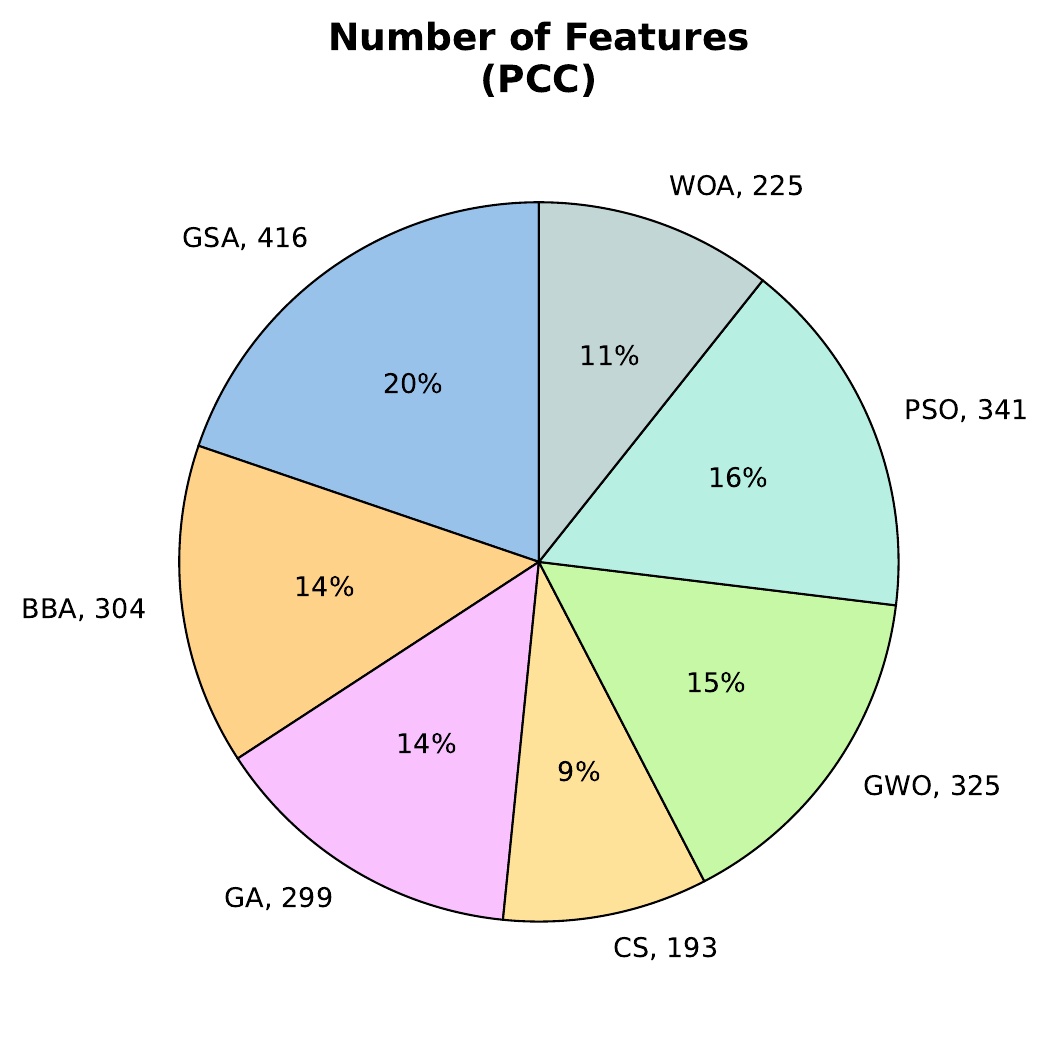}
         \caption{Mean feature count for PCC}
         \label{fig:PCC_feat}
     \end{subfigure}
     \hfill
     \begin{subfigure}[b]{0.32\textwidth}
         \centering
         \includegraphics[width=\textwidth]{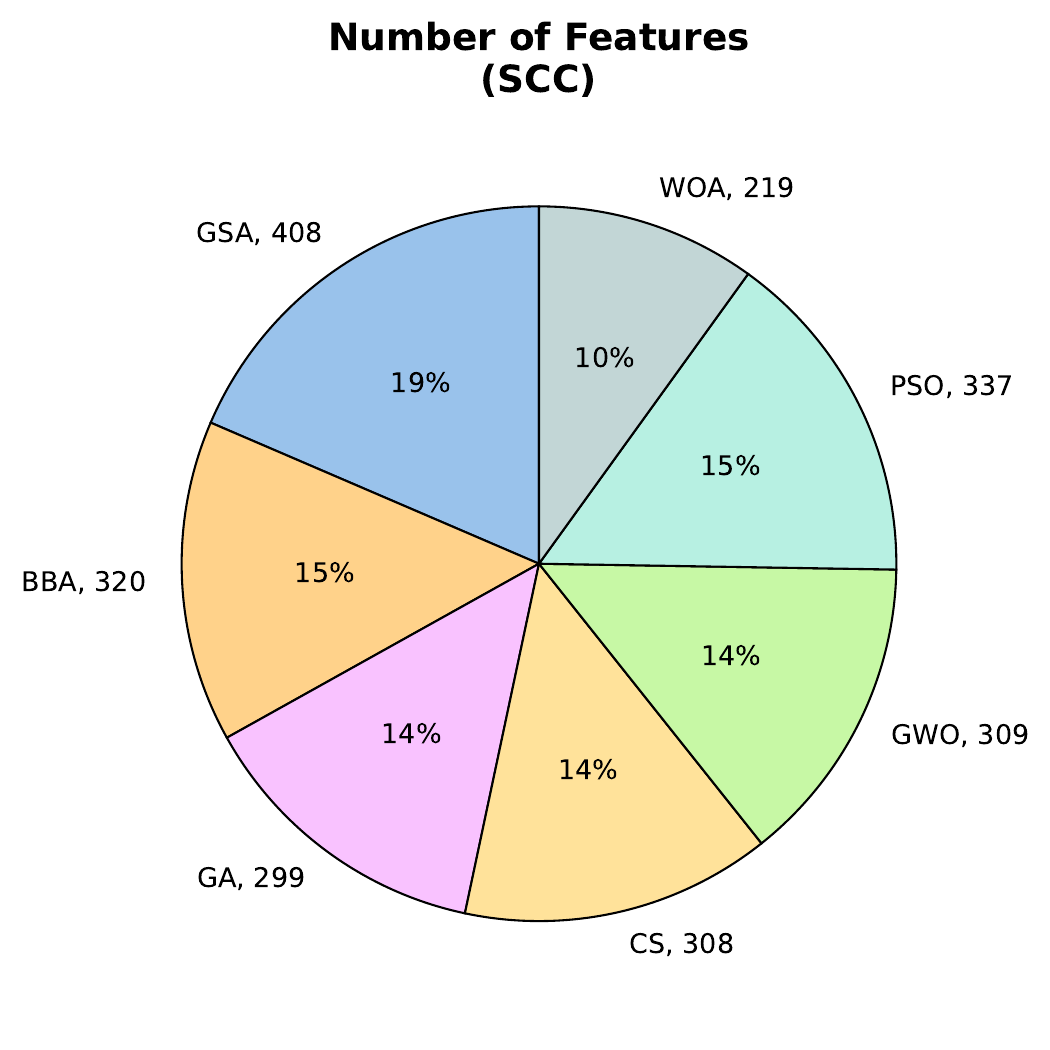}
         \caption{Mean feature count for SCC}
         \label{fig:SCC_feat}
     \end{subfigure}
     \hfill
     \begin{subfigure}[b]{0.32\textwidth}
         \centering
         \includegraphics[width=\textwidth]{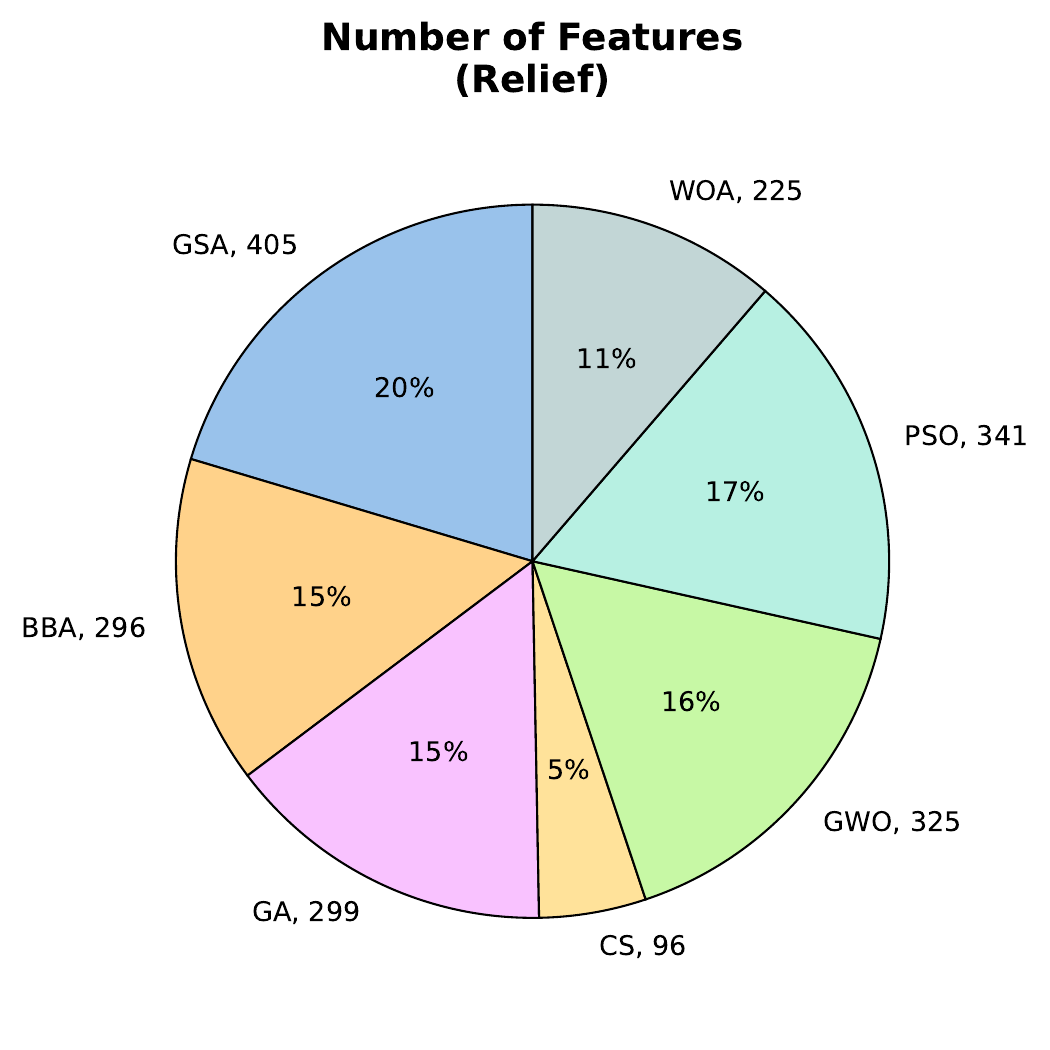}
         \caption{Mean feature count for Relief}
         \label{fig:Relief_feat}
     \end{subfigure}
     \hfill
     \caption{Feature analysis of various nature-inspired algorithms}
    \label{fig:FeatureCount}
\end{figure}

\begin{figure}
     \centering
     \begin{subfigure}[b]{0.75\textwidth}
         \centering
         \includegraphics[width=\textwidth]{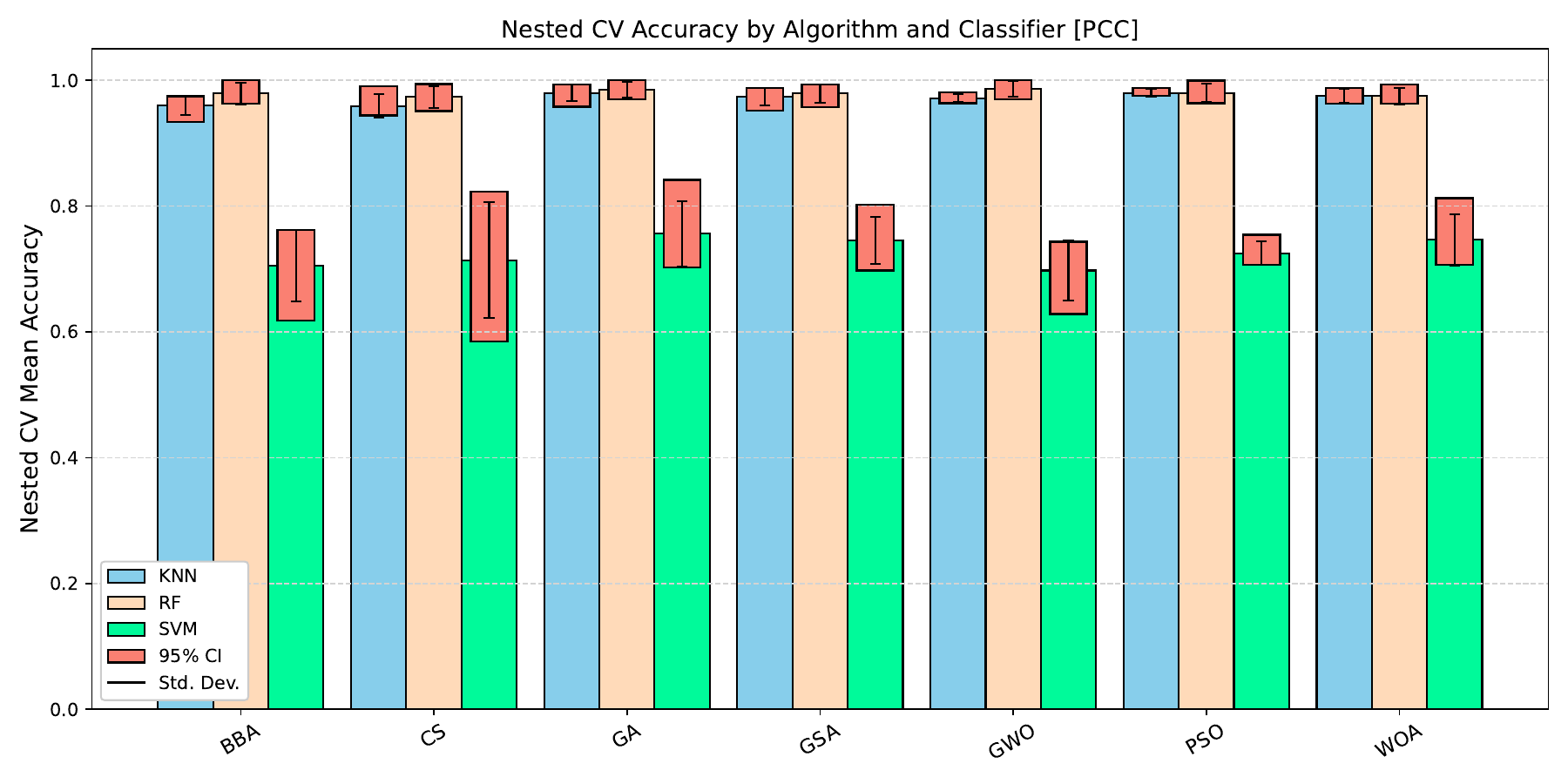}
         \caption{Nested CV accuracy analysis using PCC}
     \end{subfigure}
     \hfill
     \begin{subfigure}[b]{0.75\textwidth}
         \centering
         \includegraphics[width=\textwidth]{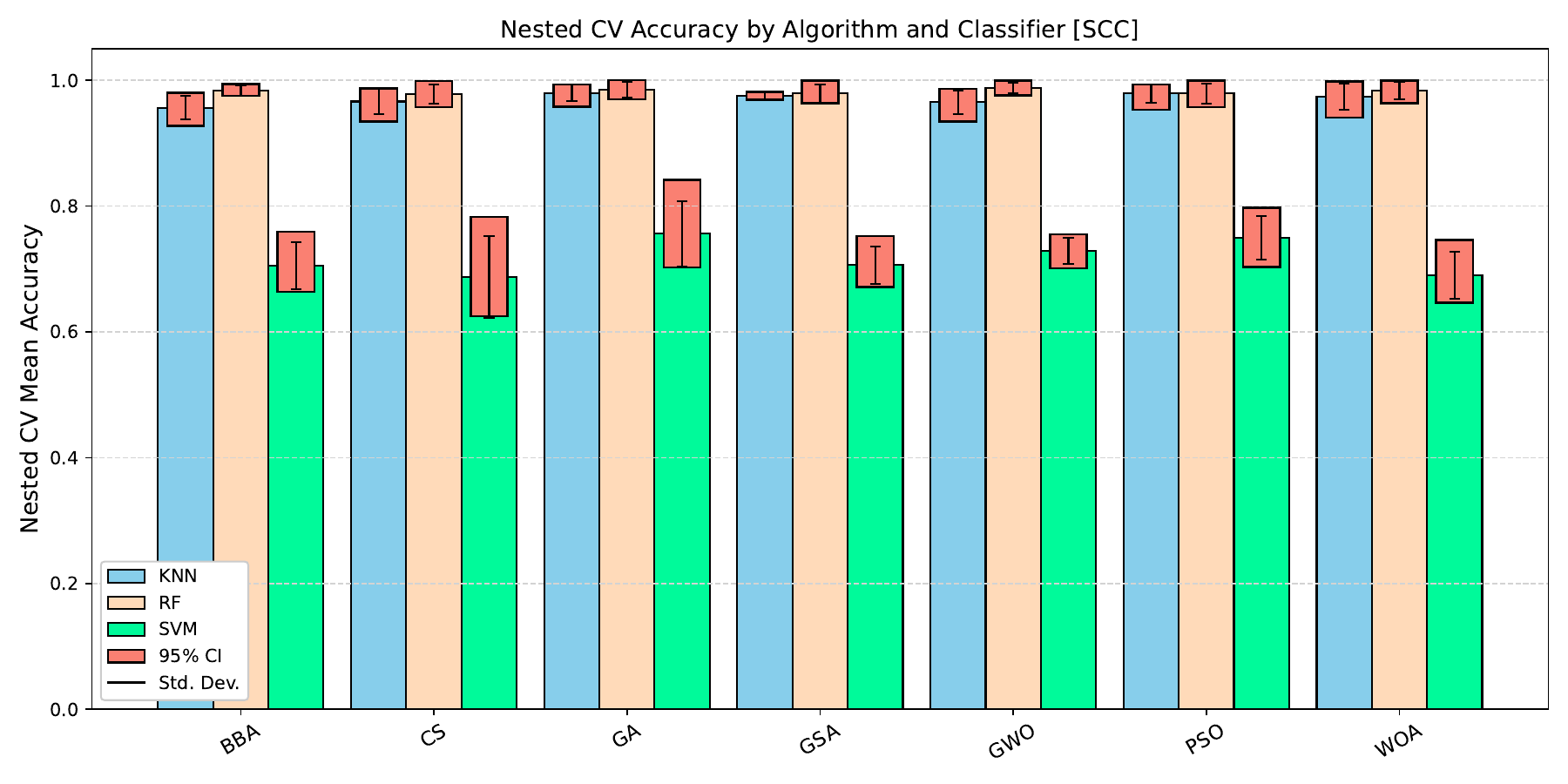}
         \caption{Nested CV accuracy analysis using SCC}
     \end{subfigure}
     \hfill
     \begin{subfigure}[b]{0.75\textwidth}
         \centering
         \includegraphics[width=\textwidth]{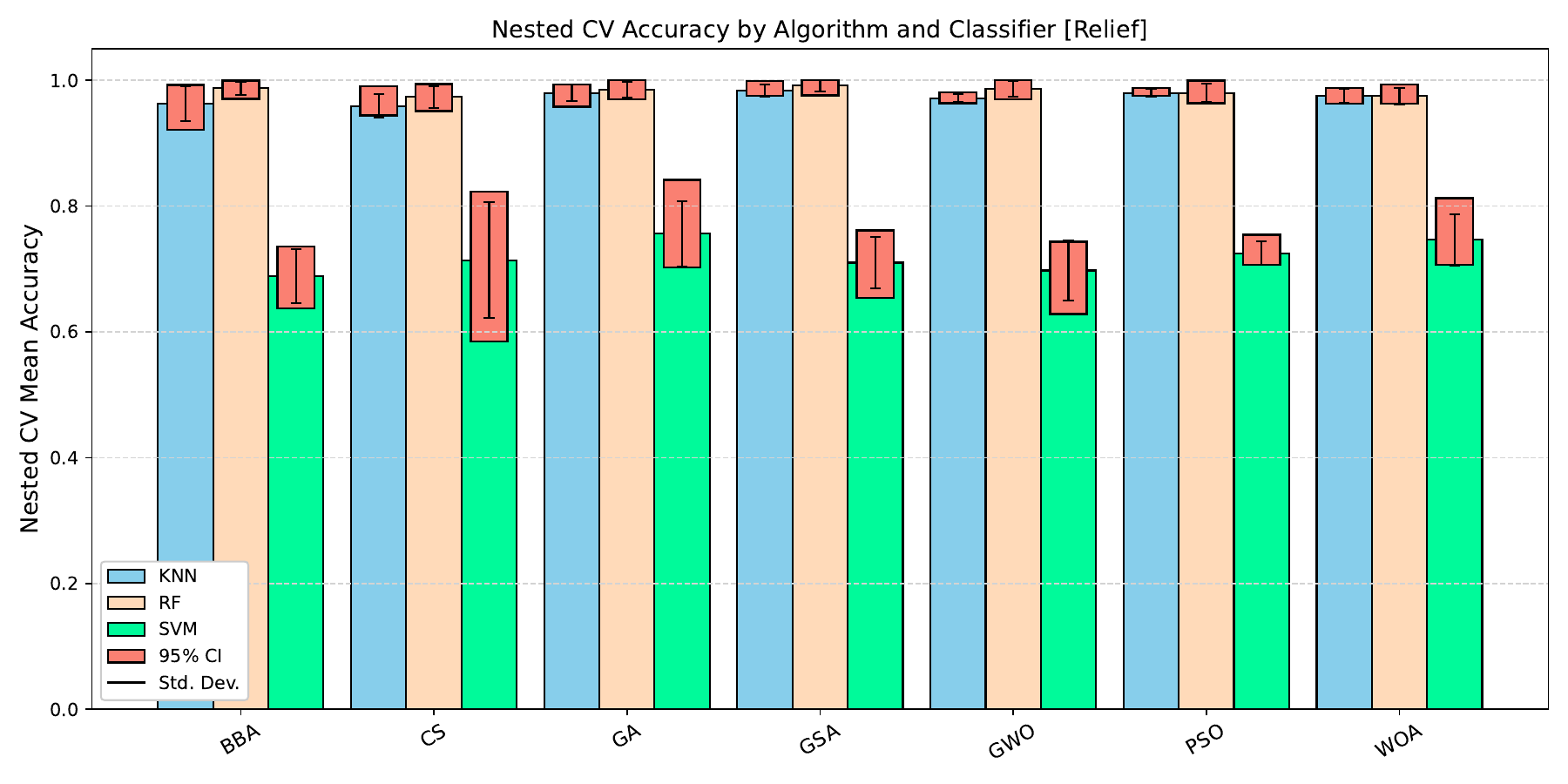}
         \caption{Nested CV accuracy analysis using Relief}
     \end{subfigure}
        \caption{Nested Cross-Validation Accuracy Analysis with 95\% Confidence Interval and Standard Deviation}
        \label{fig:NestedCV}
\end{figure}

\begin{figure}
     \centering
     \begin{subfigure}[b]{\textwidth}
         \centering
         \includegraphics[width=\textwidth]{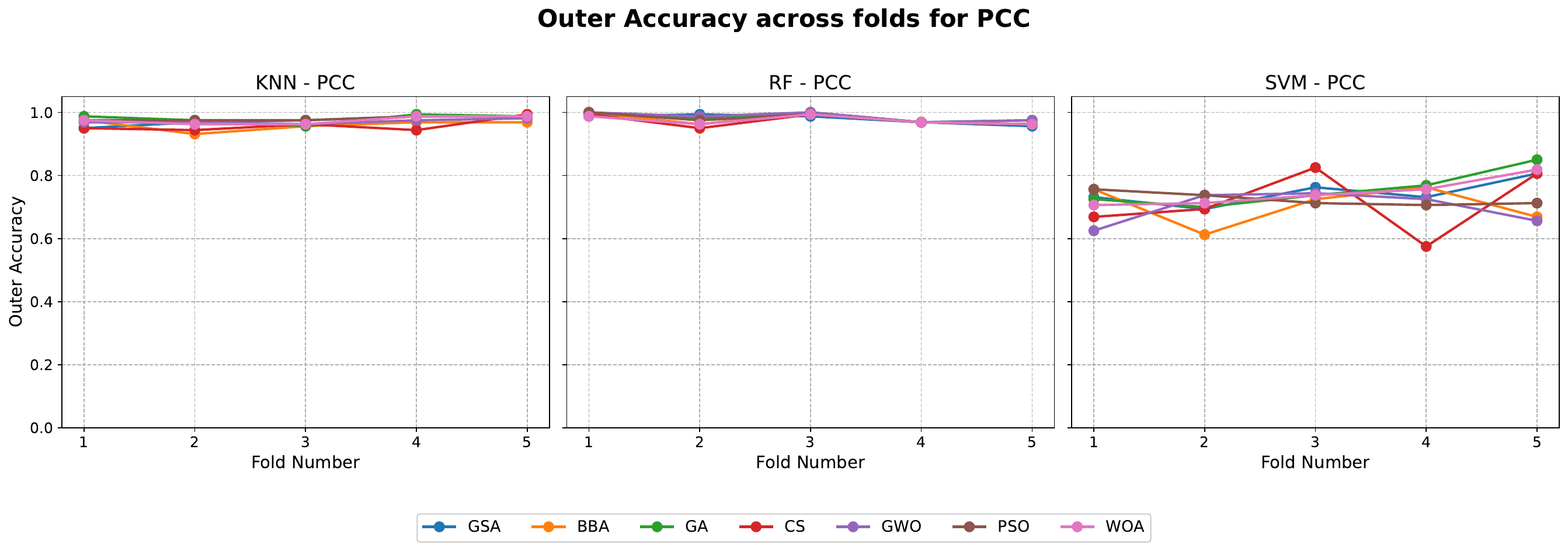}
         \caption{Pearson Correlation Coefficient}
     \end{subfigure}
     \begin{subfigure}[b]{\textwidth}
         \centering
         \includegraphics[width=\textwidth]{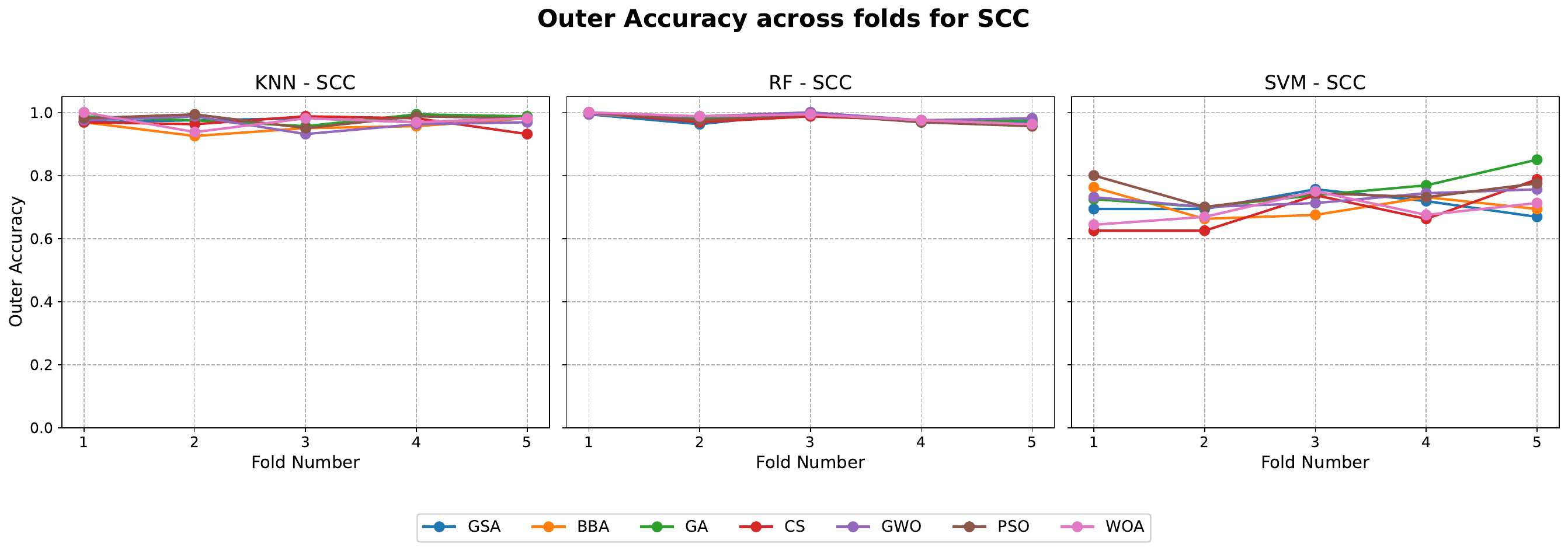}
         \caption{Spearman Correlation Coefficient}
     \end{subfigure}
     \begin{subfigure}[b]{\textwidth}
         \centering
         \includegraphics[width=\textwidth]{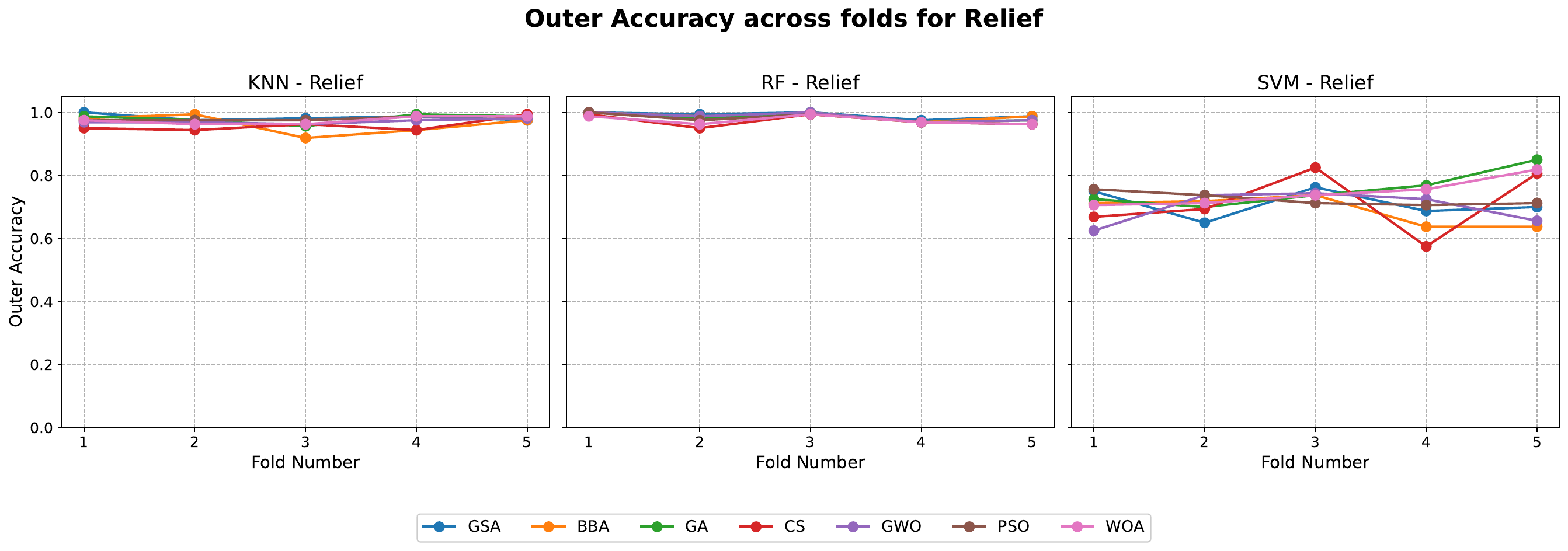}
         \caption{Relief Ranking Coefficient}
     \end{subfigure}
        \caption{Analysis of outer accuracy across the folds during training}
        \label{fig:OuterAccuracy}
\end{figure}

\begin{figure}
     \centering
     \begin{subfigure}[b]{0.49\textwidth}
         \centering
         \includegraphics[width=\textwidth]{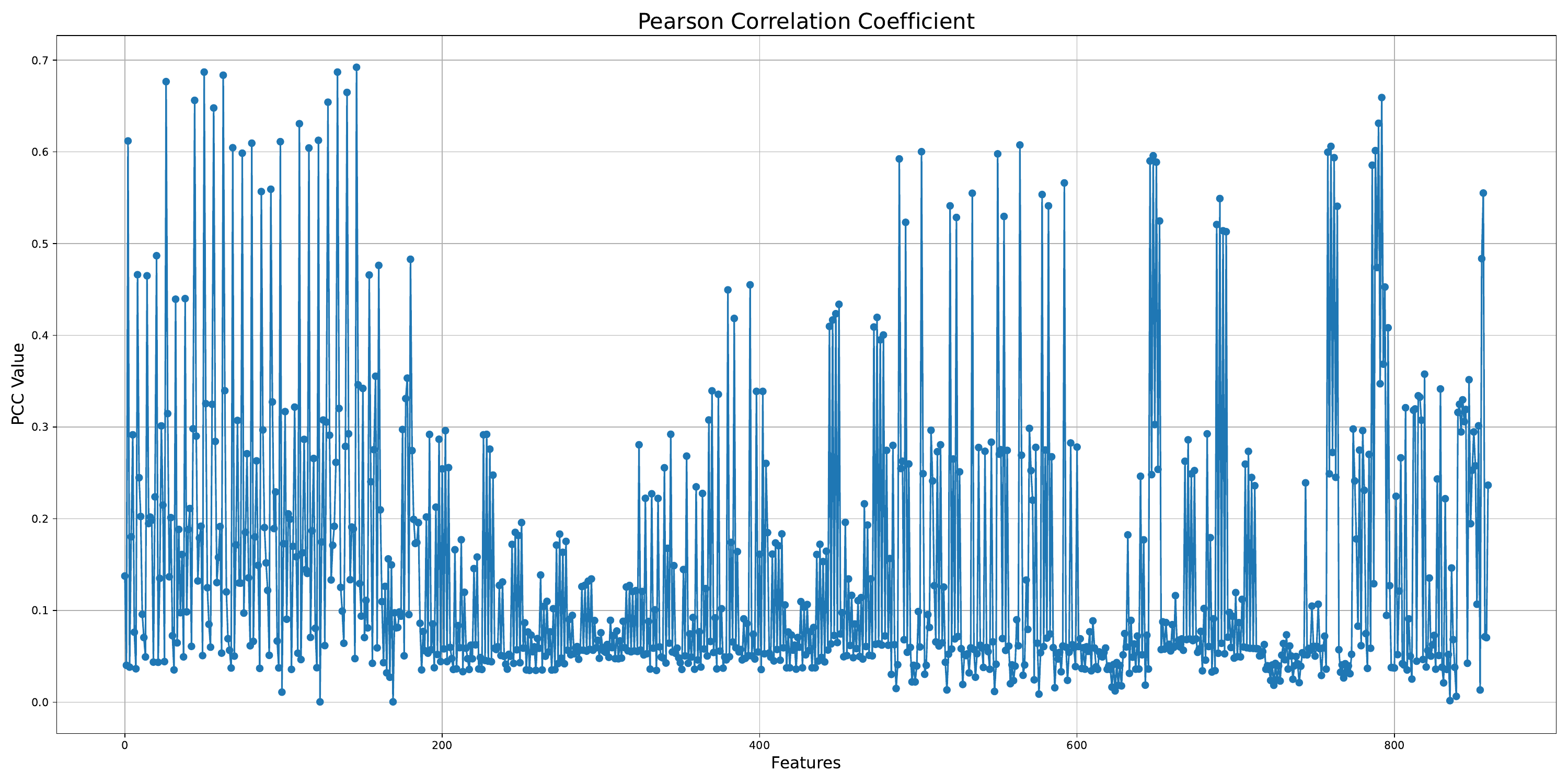}
         \caption{Pearson Correlation Coefficient}
     \end{subfigure}
     \hfill
     \begin{subfigure}[b]{0.49\textwidth}
         \centering
         \includegraphics[width=\textwidth]{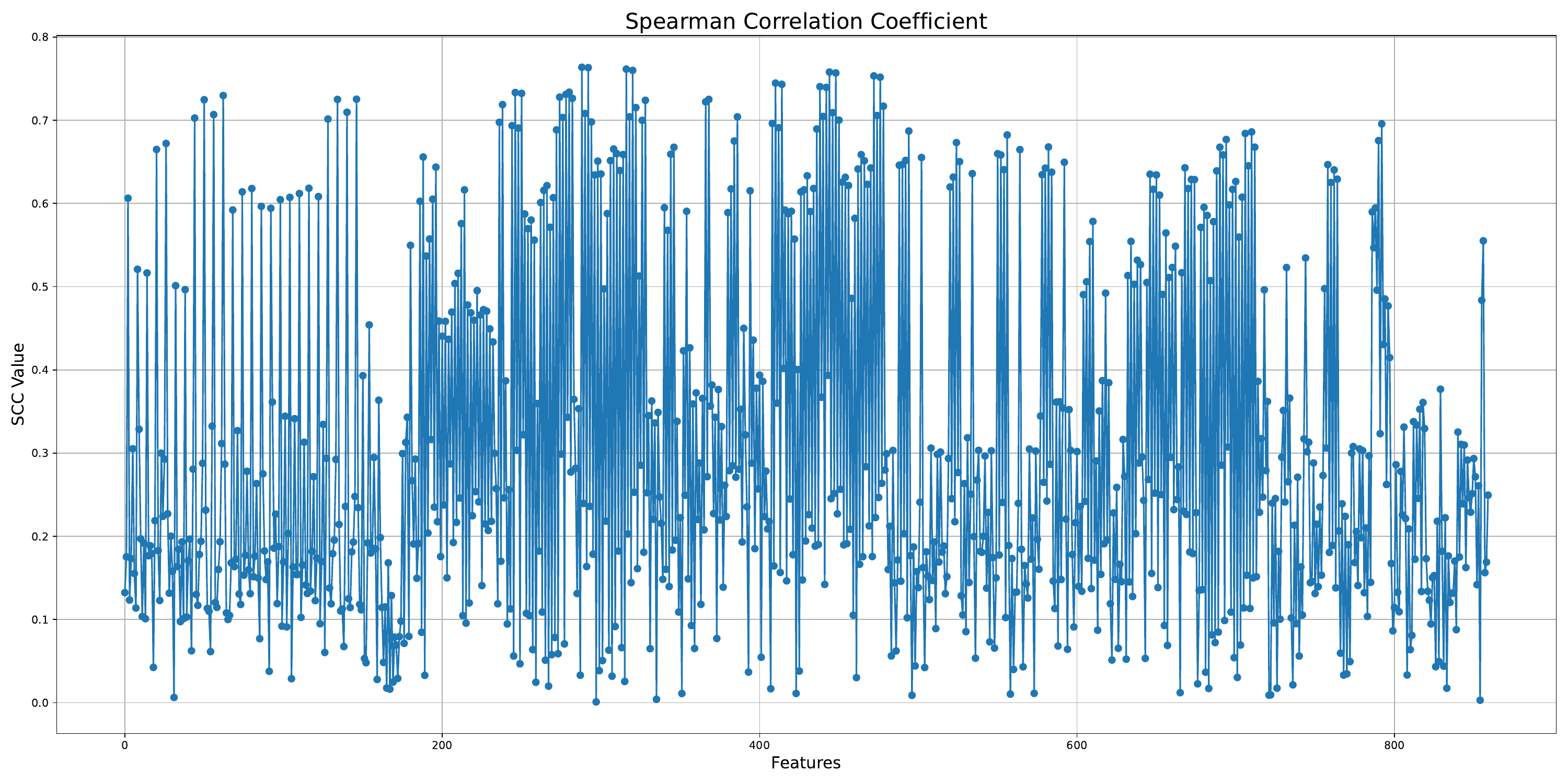}
         \caption{Spearman Correlation Coefficient}
     \end{subfigure}
     \hfill
     \begin{subfigure}[b]{0.49\textwidth}
         \centering
         \includegraphics[width=\textwidth]{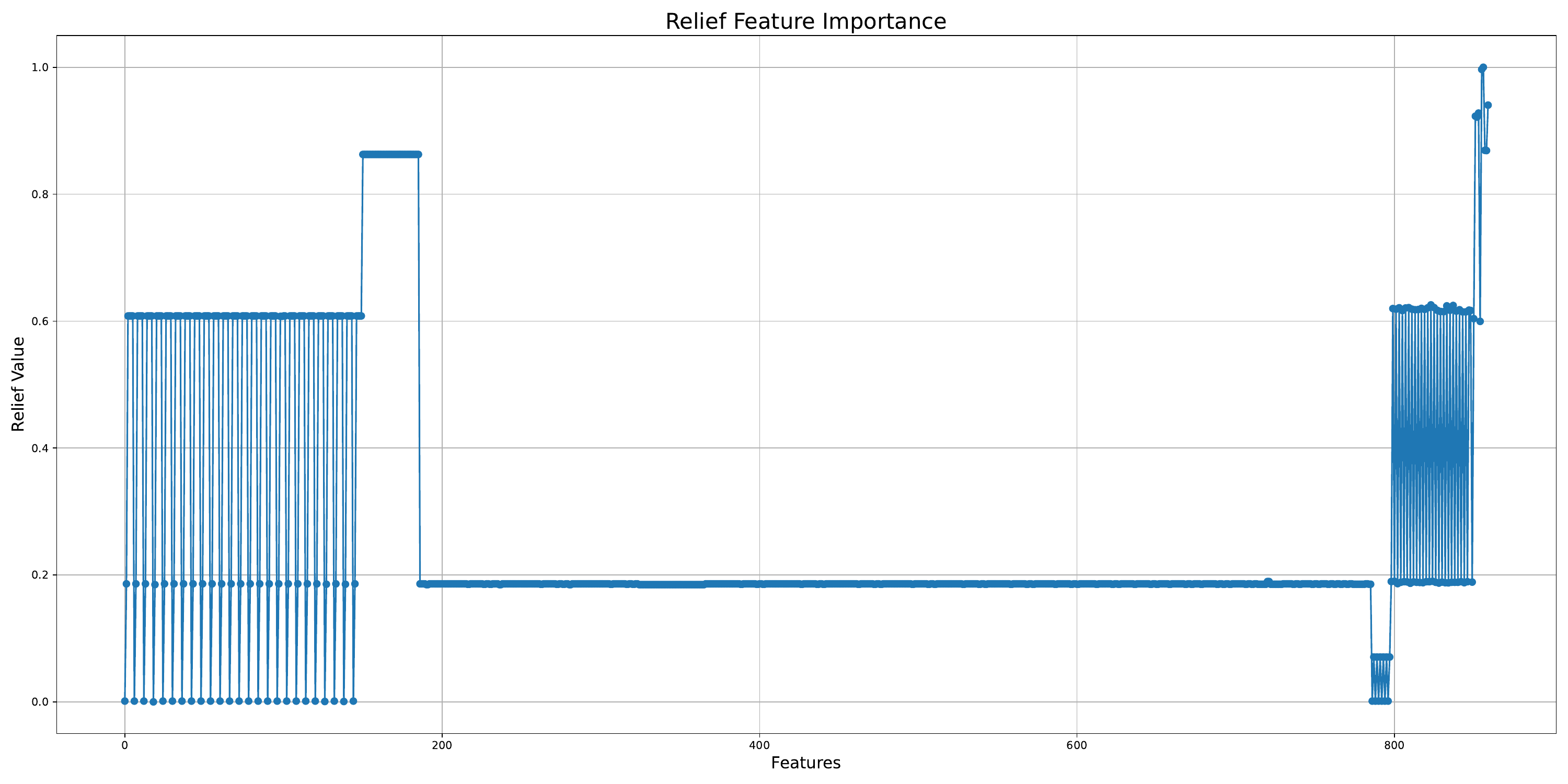}
         \caption{Relief Ranking Coefficient}
     \end{subfigure}
        \caption{Feature Importance Values}
        \label{fig:FeatureImportance}
\end{figure}

The average test accuracy analysis is visually represented using the histogram shown in Fig.~\ref{fig:accuracy}, where different colors correspond to various nature-inspired optimization algorithms. The histogram is divided into two sections, highlighting the performances of different classifiers and feature ranking coefficients utilized during evaluation. It is evident from the histogram that the SVM consistently performs the worst across all combinations of ranking coefficients and classifiers, whereas the Relief ranking coefficient yields the best results overall. It is important to note that these accuracies represent average values, hence, for a more comprehensive understanding of model generalizability, one must also consider the results presented in Fig.~\ref{fig:NestedCV} and Fig.~\ref{fig:OuterAccuracy}. Furthermore, the pie charts depicted in Fig.~\ref{fig:FeatureCount} illustrate the mean number of features selected for each combination, while Fig.~\ref{fig:FeatureImportance} represents the feature importance scores used for initial particle selection.

Fig.~\ref{fig:NestedCV} depicts the nested cross-validation accuracy trends achieved by various nature-inspired optimization algorithms across the feature ranking coefficients. Each subplot corresponds to a different ranking coefficient, with multiple colored lines representing distinct classifier algorithms. Among all optimizers, the GSA demonstrated the smallest standard deviation and narrowest confidence intervals in accuracy, particularly when paired with the Relief ranking coefficient and the RF classifier. Both KNN and RF classifiers showed minimal fluctuations in accuracy over different data splits for most optimizers, indicating a robust feature selection pipeline and strong classifier generalization. Additionally, the GWO combined with the SCC ranking coefficient and RF classifier also exhibited low variability and tight accuracy intervals. These observations suggest multiple viable model configurations, allowing practitioners to select models based on desired characteristics such as accuracy, stability, or computational efficiency. Fig.~\ref{fig:OuterAccuracy} visualizes the outer accuracy performance statistics across the folds for each combination of nature-inspired optimization algorithms and classifier, across all feature selection methods. Each subplot corresponds to a specific ranking approach, providing a direct comparative view of their impact. Across all plots, the RF classifier consistently. Notably, the accuracy remains high for virtually all nature algorithms, demonstrating the effectiveness of the pipeline's stratified cross-validation and feature optimization.

\begin{figure}
     \centering
     \begin{subfigure}[b]{\textwidth}
         \centering
         \includegraphics[width=\textwidth]{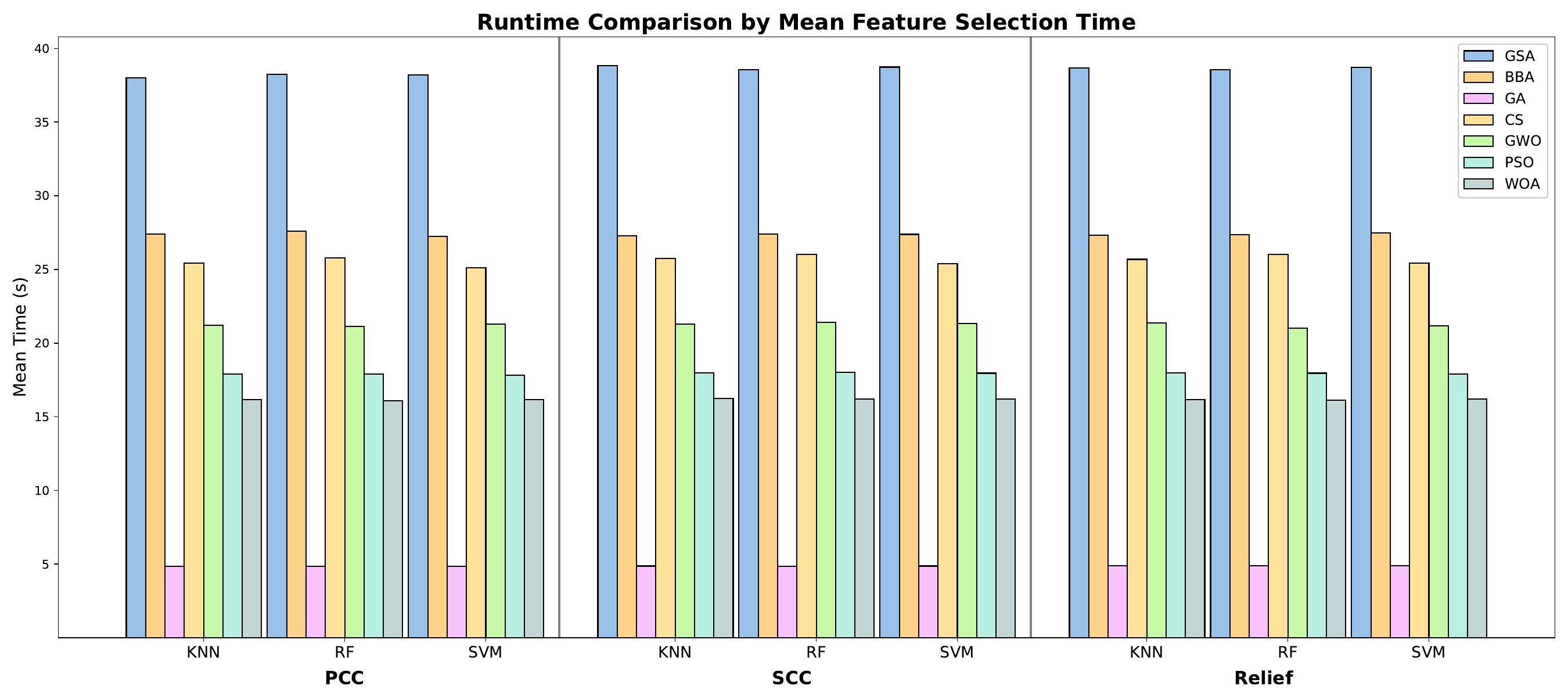}
         \caption{Runtime comparison of mean feature selection time}
     \end{subfigure}
     \begin{subfigure}[b]{\textwidth}
         \centering
         \includegraphics[width=\textwidth]{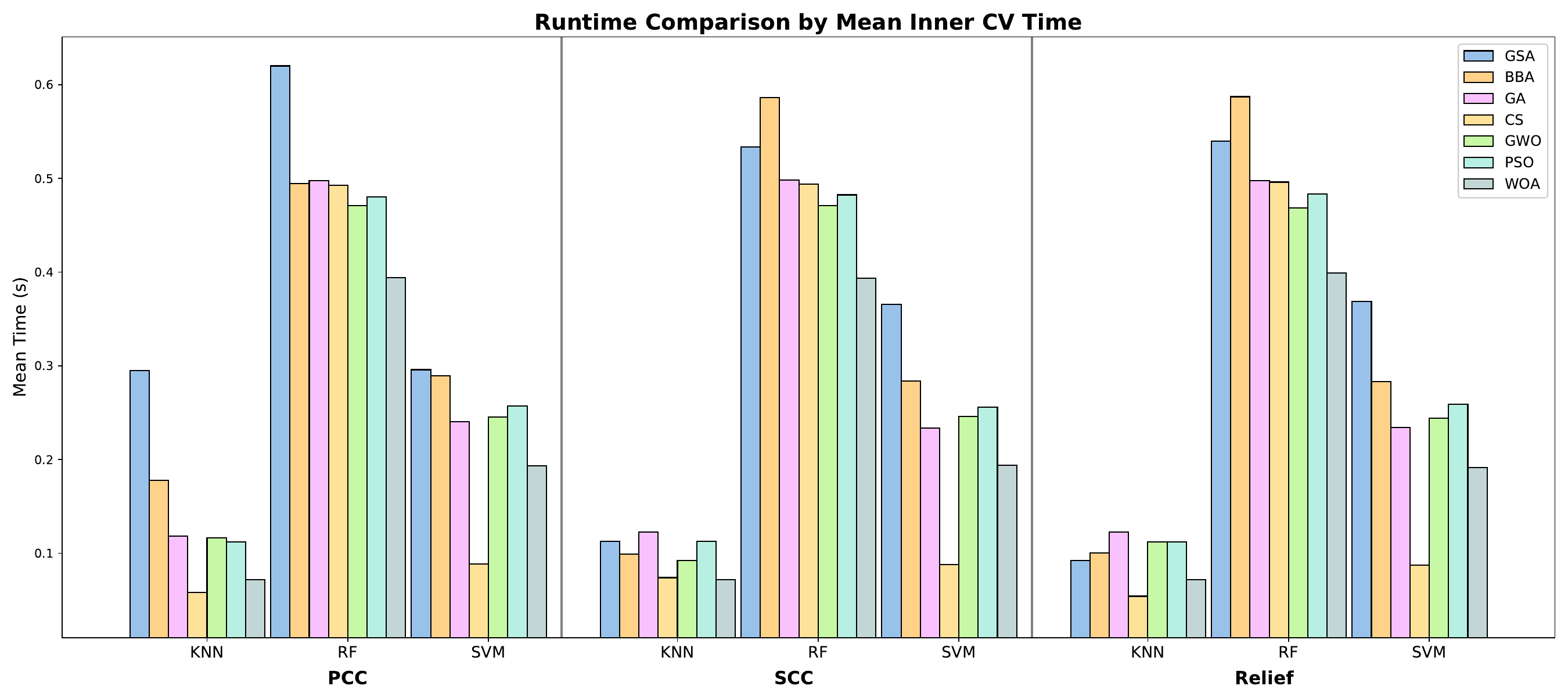}
         \caption{Runtime comparison of mean inner cross-validation time}
     \end{subfigure}
        \caption{Runtime comparisons of various combinations of feature extraction and classification algorithms}
        \label{fig:Runtime}
\end{figure}

\begin{table}
    \centering
    \caption{Best Results obtained out of all the test runs}
    \label{tab:FinalResults}
    \resizebox{\columnwidth}{!}{
    \begin{tabular}{|c|c|c|c|c|} \hline 
         \textbf{Combination of Algorithms} &   \textbf{Accuracy} & \textbf{Time (s)} & \textbf{Avg. Features} & \textbf{Feature Reduction} \\ \hline 
         Relief + GSA + RF & \textbf{100}\% & $\approx 39$ & 380 & 69.81\% \\ \hline
         Relief + CS + RF & 96.875\% & $\approx 27$ & \textbf{4} & \textbf{99.68\%} \\\hline
         SCC + GA + KNN & 98.75\% & $\mathbf{\approx 5}$ & 399 & 68.31\% \\\hline
    \end{tabular}}
\end{table}

\subsection{Comparison with existing methods}
Comparisons are done with state-of-the-art methods, namely, Bimodal feature analysis with deep learning for autism spectrum disorder detection~\cite{colonnese2024bimodal}, Machine Learning Approach for Identification of Autism Spectrum Disorder from Video Using OpenPose~\cite{kalam2024machine}, and Autism Spectrum Disorder Detection Using Skeleton-Based Body Movement Analysis via Dual-Stream Deep Learning~\cite{shin2025autism}. The results presented in Table~\ref{tab:PerformanceComparison} clearly demonstrate the efficacy of the proposed Relief + GSA + RF framework in comparison with existing state-of-the-art approaches for ASD detection. Notably, while ConcatNet~\cite{colonnese2024bimodal}, among the bimodal fusion architectures, attained a strong performance with an accuracy of $0.986$, the proposed method surpassed this benchmark by achieving an accuracy of $0.991$, thereby establishing a new performance ceiling within this domain. Unimodal video-based methods, including Random Forest with PCA~\cite{kalam2024machine}, yielded notably lower accuracy ($0.950$) and exhibited instability as reflected by a larger standard deviation ($0.1136$). Similarly, skeleton-based dual-stream deep learning methods~\cite{shin2025autism}, encompassing both convolutional and transformer backbones, exhibited performance clustered within the $0.932-0.954$ range. However, their robustness varied distinctly, with models such as ViT-B yielding highly unstable outputs (standard deviation $0.8300$), raising concerns about their generalizability. An additional distinguishing factor of the proposed method lies in its computational efficiency. Specifically, Relief + GSA + RF achieved the lowest recorded training time ($39.10$ sec), substantially outperforming the training times of the neural architectures from~\cite{colonnese2024bimodal}, which ranged from $47.45$ sec to $80.72$ sec. Collectively, these findings underscore that the integration of targeted feature reduction strategies with ensemble learning provides a compelling alternative to purely deep learning-based solutions, offering an improved balance between accuracy, robustness, and computational practicality.

\begin{table}
    \centering
    \caption{Comparative Performance Analysis with State-Of-The-Art}
    \label{tab:PerformanceComparison}
    \begin{tabular}{|l|c|c|c|}
        \hline
        \textbf{Model} & \textbf{Accuracy} & \textbf{Std. Dev.} & \textbf{Train Time (s)}\\\hline
        LandNet~\cite{colonnese2024bimodal} & 0.974 & 0.0140 & 80.72 \\\hline
        AngleNet~\cite{colonnese2024bimodal} & 0.975 & 0.0130 & 53.32 \\\hline
        ConcatNet~\cite{colonnese2024bimodal} & 0.986 & 0.0120 & 47.45 \\\hline
        Random Forest + PCA~\cite{kalam2024machine} & 0.950 & 0.1136 & -- \\\hline
        ViT-B (Transformer)~\cite{shin2025autism} & 0.948 & 0.8300 & -- \\\hline
        ConvNeXt-Base (CNN)~\cite{shin2025autism} & 0.950 & 0.0100 & -- \\\hline
        Swin-V2-B (Transformer)~\cite{shin2025autism} & 0.932 & 0.0083 & -- \\\hline
        ShuffleNetV2-X2 (CNN)~\cite{shin2025autism} & 0.954 & 0.0114 & -- \\\hline
        ResNet-152 (CNN)~\cite{shin2025autism} & 0.940 & \textbf{0.0071} & -- \\\hline
        MaxViT-T (CNN + Transformer)~\cite{shin2025autism} & 0.954 & 0.0182 & -- \\\hline
        \textbf{Relief + GSA + RF (ours)} & \textbf{0.991} & 0.0100 & \textbf{39.10} \\\hline
    \end{tabular}
\end{table}

\section{Conclusion and discussions}\label{sec:conclusion}
The increasing global prevalence of ASD underscores the urgent need for reliable, efficient, and accurate methods for its early prediction and intervention. This study addresses this pressing problem using a 3D walking video dataset by integrating correlation-based ranking coefficients, nature-inspired optimization algorithms for feature selection, and supervised machine learning classifiers. Unlike prior approaches, the proposed methodology demonstrates a superior balance between accuracy, feature reduction, and computational efficiency, establishing new performance benchmarks in ASD detection.  

Among all evaluated combinations, two distinct configurations emerged as particularly effective. The setup involving the Relief ranking coefficient, CS for feature selection, and RF as the classifier achieved an accuracy of $96.875\%$ with only $4$ features, corresponding to a remarkable reduction of $99.68\%$. This makes it highly suitable for scenarios prioritizing model compactness and resource efficiency. On the other hand, the integration of the Relief coefficient, GSA, and RF achieved perfect accuracy ($100\%$) with a $69.81\%$ feature reduction, thereby setting a new performance ceiling for ASD detection with state-of-the-art efficiency (training time $\approx 39$s). Furthermore, comparative analysis with existing deep learning-based methods confirms that the proposed Relief+GSA+RF framework not only surpasses them in terms of accuracy but also provides greater robustness with significantly lower computational burden. The experimental findings highlight the potential of combining correlation-driven initialization, feature selection via nature-inspired algorithms, and classification strategies to address complex neurodevelopmental diagnostic tasks. Future extensions of this work should include large-scale datasets with greater variability to validate generalizability and to further enhance adaptability and real-world deployment. In particular, modifications to the GSA for alternative distance functions and multi-dimensional optimization may expand its applicability in broader feature extraction tasks.  

Overall, the proposed methodology demonstrates that robust feature selection, when coupled with an efficient classification pipeline, can significantly improve both prediction accuracy and computational feasibility for ASD detection from video data. With perfect diagnostic accuracy achieved in the best configuration, the model shows practical potential for integration into healthcare workflows, thereby contributing to early identification and improved treatment planning for individuals with ASD. Continuous research and refinement remain essential to ensure scalability across diverse populations and to enable real-time deployment through computer vision-based systems in clinical practice.

\section*{Declarations}\label{Declarations}
\subsection*{Ethical Approval}
This study involved secondary analysis of a publicly available dataset and did not require additional ethical approval.

\subsection*{Consent to Participate}
Not applicable.

\subsection*{Consent to Publish}
Not applicable.

\subsection*{Data Availability Statement}
We used the publicly available dataset~\cite{1}, "\textit{Three-Dimensional Dataset Combining Gait and Full Body Movement of Children with Autism Spectrum Disorders Collected by Kinect v2 Camera Dataset}," which can be accessed from \href{https://datadryad.org/dataset/doi:10.5061/dryad.s7h44j150}{https://datadryad.org/dataset/doi:10.5061/dryad.s7h44j150}.

\subsection*{Authors Contributions}
\textbf{Aneesh Panchal:} Writing - original draft, Methodology, Investigation, Conceptualization. \textbf{Kainat Khan:} Writing - original draft, Supervision. \textbf{Rahul Katarya:} Writing - review \& editing, Supervision.

\subsection*{Funding}
This research received no specific grant from any funding agency in the public, commercial, or not-for-profit sectors.

\subsection*{Competing Interests}
The authors declare that they have no known competing interests.

\bibliographystyle{unsrt}
\bibliography{bibliography}

\section*{Appendix}
\subsection*{Extended Results and Explanations}
Table~\ref{tab:comprehensive_results} summarizes the single-case test accuracy, precision, recall, and F1-score for various combinations of nature-inspired optimization algorithms (optimizers), classifiers, and feature ranking coefficients. Overall, the results indicate that SVM consistently performs the worst among the classifiers evaluated.

\begin{table}[H]
\centering
\caption{Test Accuracy (Acc.), Precision (Prec.), Recall (Rec.), and F1-Score (F1) (\%)}
\label{tab:comprehensive_results}
\resizebox{\textwidth}{!}{
\begin{tabular}{|c|c|c|c|c|c|c|c|c|c|c|c|c|c|}
\hline
\multirow{2}{*}{\textbf{Ranking}} & \multirow{2}{*}{\textbf{Optimizer}} & \multicolumn{4}{c|}{\textbf{KNN}} & \multicolumn{4}{c|}{\textbf{RF}} & \multicolumn{4}{c|}{\textbf{SVM}} \\
\cline{3-14}
& & \textbf{Acc.} & \textbf{Prec.} & \textbf{Rec.} & \textbf{F1} & \textbf{Acc.} & \textbf{Prec.} & \textbf{Rec.} & \textbf{F1} & \textbf{Acc.} & \textbf{Prec.} & \textbf{Rec.} & \textbf{F1} \\
\hline
\multirow{7}{*}{PCC} & GSA & 98.75 & 98.75 & 98.75 & 98.75 & 95.62 & 95.69 & 95.62 & 95.62 & 80.62 & 86.04 & 80.62 & 79.87 \\
 & BBA & 96.88 & 96.94 & 96.88 & 96.87 & 96.25 & 96.28 & 96.25 & 96.25 & 66.88 & 66.90 & 66.88 & 66.86 \\
 & GA & 98.75 & 98.78 & 98.75 & 98.75 & \textbf{97.50} & \textbf{97.50} & \textbf{97.50} & \textbf{97.50} & \textbf{85.00} & \textbf{87.86} & \textbf{85.00} & \textbf{84.71} \\
 & CS & \textbf{99.38} & \textbf{99.38} & \textbf{99.38} & \textbf{99.37} & 96.25 & 96.28 & 96.25 & 96.25 & 80.62 & 81.46 & 80.62 & 80.50 \\
 & GWO & 98.12 & 98.19 & 98.12 & 98.12 & 97.50 & 97.53 & 97.50 & 97.50 & 65.62 & 66.56 & 65.62 & 65.13 \\
 & PSO & 98.75 & 98.78 & 98.75 & 98.75 & 96.25 & 96.28 & 96.25 & 96.25 & 71.25 & 78.33 & 71.25 & 69.33 \\
 & WOA & 98.75 & 98.75 & 98.75 & 98.75 & 96.25 & 96.28 & 96.25 & 96.25 & 81.88 & 85.97 & 81.88 & 81.34 \\
\hline
\multirow{7}{*}{SCC} & GSA & 98.12 & 98.13 & 98.12 & 98.12 & 96.88 & 96.88 & 96.88 & 96.87 & 66.88 & 78.43 & 66.88 & 63.13 \\
 & BBA & 98.12 & 98.19 & 98.12 & 98.12 & 97.50 & 97.62 & 97.50 & 97.50 & 69.38 & 69.62 & 69.38 & 69.28 \\
 & GA & \textbf{98.75} & \textbf{98.78} & \textbf{98.75} & \textbf{98.75} & 97.50 & 97.50 & 97.50 & 97.50 & \textbf{85.00} & \textbf{87.86} & \textbf{85.00} & \textbf{84.71} \\
 & CS & 93.12 & 93.46 & 93.12 & 93.11 & 95.62 & 95.69 & 95.62 & 95.62 & 78.75 & 83.45 & 78.75 & 77.98 \\
 & GWO & 96.88 & 97.06 & 96.88 & 96.87 & \textbf{98.12} & \textbf{98.19} & \textbf{98.12} & \textbf{98.12} & 75.62 & 80.88 & 75.62 & 74.54 \\
 & PSO & 98.12 & 98.19 & 98.12 & 98.12 & 95.62 & 95.69 & 95.62 & 95.62 & 77.50 & 82.74 & 77.50 & 76.56 \\
 & WOA & 98.12 & 98.13 & 98.12 & 98.12 & 96.25 & 96.28 & 96.25 & 96.25 & 71.25 & 79.34 & 71.25 & 69.12 \\
\hline
\multirow{7}{*}{Relief} & GSA & 97.50 & 97.50 & 97.50 & 97.50 & \textbf{98.75} & \textbf{98.78} & \textbf{98.75} & \textbf{98.75} & 70.00 & 78.67 & 70.00 & 67.55 \\
 & BBA & 97.50 & 97.53 & 97.50 & 97.50 & \textbf{98.75} & \textbf{98.78} & \textbf{98.75} & \textbf{98.75} & 63.75 & 64.32 & 63.75 & 63.38 \\
 & GA & 98.75 & 98.78 & 98.75 & 98.75 & 97.50 & 97.50 & 97.50 & 97.50 & \textbf{85.00} & \textbf{87.86} & \textbf{85.00} & \textbf{84.71} \\
 & CS & \textbf{99.38} & \textbf{99.38} & \textbf{99.38} & \textbf{99.37} & 96.25 & 96.28 & 96.25 & 96.25 & 80.62 & 81.46 & 80.62 & 80.50 \\
 & GWO & 98.12 & 98.19 & 98.12 & 98.12 & 97.50 & 97.53 & 97.50 & 97.50 & 65.62 & 66.56 & 65.62 & 65.13 \\
 & PSO & 98.75 & 98.78 & 98.75 & 98.75 & 96.25 & 96.28 & 96.25 & 96.25 & 71.25 & 78.33 & 71.25 & 69.33 \\
 & WOA & 98.75 & 98.75 & 98.75 & 98.75 & 96.25 & 96.28 & 96.25 & 96.25 & 81.88 & 85.97 & 81.88 & 81.34 \\
\hline
\end{tabular}}
\end{table}

Table~\ref{tab:nested_cv_results} reports the nested cross-validation outcomes, including mean accuracy, standard deviation, and $95\%$ confidence intervals for the different algorithmic combinations. Consistent with the previous findings, SVM exhibits the poorest performance, characterized by low accuracy and high variability.

\begin{table}[H]
\centering
\caption{Nested Cross-Validation Results: Mean Accuracy (Acc.), Standard Deviation (SD), and Confidence Intervals (CI)}
\label{tab:nested_cv_results}
\resizebox{\textwidth}{!}{
\begin{tabular}{|c|c|c|c|c|c|c|c|c|c|c|}
\hline
\multirow{2}{*}{\textbf{Ranking}} & \multirow{2}{*}{\textbf{Optimizer}} & \multicolumn{3}{c|}{\textbf{KNN}} & \multicolumn{3}{c|}{\textbf{RF}} & \multicolumn{3}{c|}{\textbf{SVM}} \\
\cline{3-11}
& & \textbf{Acc.} & \textbf{SD} & \textbf{CI} & \textbf{Acc.} & \textbf{SD} & \textbf{CI} & \textbf{Acc.} & \textbf{SD} & \textbf{CI} \\
\hline
\multirow{7}{*}{PCC} & GSA & 97.37 & 0.02 & [95.19, 98.75] & 97.88 & 0.02 & [95.75, 99.31] & 74.50 & 0.14 & [69.75, 80.19] \\
 & BBA & 96.00 & 0.02 & [93.38, 97.44] & 97.88 & 0.03 & [96.25, 100.00] & 70.50 & 0.32 & [61.81, 76.19] \\
 & GA & \textbf{98.00} & \textbf{0.02} & \textbf{[95.81, 99.31]} & 98.50 & 0.02 & [96.94, 100.00] & \textbf{75.62} & \textbf{0.27} & \textbf{[70.25, 84.19]} \\
 & CS & 95.88 & 0.04 & [94.38, 99.06] & 97.38 & 0.03 & [95.12, 99.38] & 71.38 & 0.85 & [58.44, 82.31] \\
 & GWO & 97.12 & 0.00 & [96.31, 98.06] & \textbf{98.62} & \textbf{0.02} & \textbf{[96.94, 100.00]} & 69.75 & 0.23 & [62.81, 74.31] \\
 & PSO & 98.00 & 0.00 & [97.50, 98.75] & 98.00 & 0.02 & [96.31, 99.94] & 72.50 & 0.04 & [70.69, 75.44] \\
 & WOA & 97.50 & 0.01 & [96.25, 98.75] & 97.50 & 0.02 & [96.25, 99.31] & 74.63 & 0.16 & [70.69, 81.25] \\
\hline
\multirow{7}{*}{SCC} & GSA & 97.50 & 0.00 & [96.88, 98.12] & 97.88 & 0.02 & [96.31, 99.94] & 70.62 & 0.09 & [67.12, 75.25] \\
 & BBA & 95.62 & 0.04 & [92.75, 98.00] & 98.38 & 0.01 & [97.50, 99.38] & 70.50 & 0.14 & [66.38, 75.94] \\
 & GA & \textbf{98.00} & \textbf{0.02} & \textbf{[95.81, 99.31]} & 98.50 & 0.02 & [96.94, 100.00] & \textbf{75.62} & \textbf{0.27} & \textbf{[70.25, 84.19]} \\
 & CS & 96.63 & 0.04 & [93.44, 98.69] & 97.75 & 0.02 & [95.75, 99.88] & 68.75 & 0.42 & [62.50, 78.25] \\
 & GWO & 96.50 & 0.04 & [93.44, 98.62] & \textbf{98.75} & \textbf{0.01} & \textbf{[97.56, 99.94]} & 72.88 & 0.04 & [70.12, 75.50] \\
 & PSO & 97.88 & 0.02 & [95.31, 99.31] & 97.88 & 0.03 & [95.75, 99.94] & 75.00 & 0.12 & [70.31, 79.75] \\
 & WOA & 97.38 & 0.04 & [94.06, 99.81] & 98.38 & 0.02 & [96.38, 99.94] & 69.00 & 0.14 & [64.62, 74.62] \\
\hline
\multirow{7}{*}{Relief} & GSA & \textbf{98.37} & \textbf{0.01} & \textbf{[97.50, 99.88]} & \textbf{99.12} & \textbf{0.01} & \textbf{[97.62, 100.00]} & 71.00 & 0.17 & [65.38, 76.12] \\
 & BBA & 96.25 & 0.08 & [92.12, 99.25] & 98.75 & 0.01 & [97.06, 99.94] & 68.88 & 0.18 & [63.75, 73.56] \\
 & GA & 98.00 & 0.02 & [95.81, 99.31] & 98.50 & 0.02 & [96.94, 100.00] & \textbf{75.62} & \textbf{0.27} & \textbf{[70.25, 84.19]} \\
 & CS & 95.88 & 0.04 & [94.38, 99.06] & 97.38 & 0.03 & [95.12, 99.38] & 71.38 & 0.85 & [58.44, 82.31] \\
 & GWO & 97.12 & 0.00 & [96.31, 98.06] & 98.62 & 0.02 & [96.94, 100.00] & 69.75 & 0.23 & [62.81, 74.31] \\
 & PSO & 98.00 & 0.00 & [97.50, 98.75] & 98.00 & 0.02 & [96.31, 99.94] & 72.50 & 0.04 & [70.69, 75.44] \\
 & WOA & 97.50 & 0.01 & [96.25, 98.75] & 97.50 & 0.02 & [96.25, 99.31] & 74.63 & 0.16 & [70.69, 81.25] \\
\hline
\end{tabular}}
\end{table}

Table~\ref{tab:outer_cv_results} presents the results of outer cross-validation across multiple folds using the same combinations, further demonstrating the superior performance of KNN and RF classifiers over the SVM approach.

\begin{table}[H]
\centering
\caption{Outer Cross-Validation Accuracy (\%) over Folds \{F1, F2, F3, F4, F5\}}
\label{tab:outer_cv_results}
\resizebox{\textwidth}{!}{
\begin{tabular}{|c|c|c|c|c|c|c|c|c|c|c|c|c|c|c|c|c|}
\hline
\multirow{2}{*}{\textbf{Ranking}} & \multirow{2}{*}{\textbf{Optimizer}} & \multicolumn{5}{c|}{\textbf{KNN}} & \multicolumn{5}{c|}{\textbf{RF}} & \multicolumn{5}{c|}{\textbf{SVM}} \\
\cline{3-17}
& & \textbf{F1} & \textbf{F2} & \textbf{F3} & \textbf{F4} & \textbf{F5} & \textbf{F1} & \textbf{F2} & \textbf{F3} & \textbf{F4} & \textbf{F5} & \textbf{F1} & \textbf{F2} & \textbf{F3} & \textbf{F4} & \textbf{F5} \\
\hline
\multirow{7}{*}{PCC} & GSA & 95.00 & 96.88 & 97.50 & 98.75 & 98.75 & 98.75 & 99.38 & 98.75 & 96.88 & 95.62 & 73.12 & 69.38 & 76.25 & 73.12 & 80.62 \\
 & BBA & 97.50 & 93.12 & 95.62 & 96.88 & 96.88 & 100.00 & 96.25 & 100.00 & 96.88 & 96.25 & 75.62 & 61.25 & 72.50 & 76.25 & 66.88 \\
 & GA & 98.75 & 97.50 & 95.62 & 99.38 & 98.75 & 100.00 & 98.12 & 100.00 & 96.88 & 97.50 & 72.50 & 70.00 & 73.75 & 76.88 & 85.00 \\
 & CS & 95.00 & 94.38 & 96.25 & 94.38 & 99.38 & 99.38 & 95.00 & 99.38 & 96.88 & 96.25 & 66.88 & 69.38 & 82.50 & 57.50 & 80.62 \\
 & GWO & 96.88 & 96.88 & 96.25 & 97.50 & 98.12 & 100.00 & 98.75 & 100.00 & 96.88 & 97.50 & 62.50 & 73.75 & 74.38 & 72.50 & 65.62 \\
 & PSO & 97.50 & 97.50 & 97.50 & 98.75 & 98.75 & 100.00 & 97.50 & 99.38 & 96.88 & 96.25 & 75.62 & 73.75 & 71.25 & 70.62 & 71.25 \\
 & WOA & 97.50 & 96.25 & 96.25 & 98.75 & 98.75 & 98.75 & 96.25 & 99.38 & 96.88 & 96.25 & 70.62 & 71.25 & 73.75 & 75.62 & 81.88 \\
\hline
\multirow{7}{*}{SCC} & GSA & 96.88 & 97.50 & 98.12 & 96.88 & 98.12 & 99.38 & 96.25 & 100.00 & 96.88 & 96.88 & 69.38 & 69.38 & 75.62 & 71.88 & 66.88 \\
 & BBA & 96.88 & 92.50 & 95.00 & 95.62 & 98.12 & 99.38 & 98.12 & 99.38 & 97.50 & 97.50 & 76.25 & 66.25 & 67.50 & 73.12 & 69.38 \\
 & GA & 98.75 & 97.50 & 95.62 & 99.38 & 98.75 & 100.00 & 98.12 & 100.00 & 96.88 & 97.50 & 72.50 & 70.00 & 73.75 & 76.88 & 85.00 \\
 & CS & 96.88 & 96.25 & 98.75 & 98.12 & 93.12 & 100.00 & 96.88 & 98.75 & 97.50 & 95.62 & 62.50 & 62.50 & 73.75 & 66.25 & 78.75 \\
 & GWO & 97.50 & 98.75 & 93.12 & 96.25 & 96.88 & 99.38 & 98.75 & 100.00 & 97.50 & 98.12 & 73.12 & 70.00 & 71.25 & 74.38 & 75.62 \\
 & PSO & 98.12 & 99.38 & 95.00 & 98.75 & 98.12 & 100.00 & 97.50 & 99.38 & 96.88 & 95.62 & 80.00 & 70.00 & 74.38 & 73.12 & 77.50 \\
 & WOA & 100.00 & 93.75 & 98.12 & 96.88 & 98.12 & 100.00 & 98.75 & 99.38 & 97.50 & 96.25 & 64.38 & 66.88 & 75.00 & 67.50 & 71.25 \\
\hline
\multirow{7}{*}{Relief} & GSA & 100.00 & 97.50 & 98.12 & 98.75 & 97.50 & 100.00 & 99.38 & 100.00 & 97.50 & 98.75 & 75.00 & 65.00 & 76.25 & 68.75 & 70.00 \\
 & BBA & 98.12 & 99.38 & 91.88 & 94.38 & 97.50 & 100.00 & 98.75 & 99.38 & 96.88 & 98.75 & 71.25 & 71.88 & 73.75 & 63.75 & 63.75 \\
 & GA & 98.75 & 97.50 & 95.62 & 99.38 & 98.75 & 100.00 & 98.12 & 100.00 & 96.88 & 97.50 & 72.50 & 70.00 & 73.75 & 76.88 & 85.00 \\
 & CS & 95.00 & 94.38 & 96.25 & 94.38 & 99.38 & 99.38 & 95.00 & 99.38 & 96.88 & 96.25 & 66.88 & 69.38 & 82.50 & 57.50 & 80.62 \\
 & GWO & 96.88 & 96.88 & 96.25 & 97.50 & 98.12 & 100.00 & 98.75 & 100.00 & 96.88 & 97.50 & 62.50 & 73.75 & 74.38 & 72.50 & 65.62 \\
 & PSO & 97.50 & 97.50 & 97.50 & 98.75 & 98.75 & 100.00 & 97.50 & 99.38 & 96.88 & 96.25 & 75.62 & 73.75 & 71.25 & 70.62 & 71.25 \\
 & WOA & 97.50 & 96.25 & 96.25 & 98.75 & 98.75 & 98.75 & 96.25 & 99.38 & 96.88 & 96.25 & 70.62 & 71.25 & 73.75 & 75.62 & 81.88 \\
\hline
\end{tabular}}
\end{table}

Table~\ref{tab:runtime_performance} details the average feature selection time, cross-validation time, and the average number of features selected for each combination. These results highlight the efficiency of the GA in computational time and the effectiveness of the CS algorithm in minimizing the number of selected features.

\begin{table}[H]
\centering
\caption{Feature Selection (FS) time (in sec), Cross Validation (CV) time (in sec), and Mean number of features selected (\#Feat.)}
\label{tab:runtime_performance}
\resizebox{\textwidth}{!}{
\begin{tabular}{|c|c|c|c|c|c|c|c|c|c|c|c|}
\hline
\multirow{2}{*}{\textbf{Ranking}} & \multirow{2}{*}{\textbf{Optimizer}} & \multicolumn{3}{c|}{\textbf{KNN}} & \multicolumn{3}{c|}{\textbf{RF}} & \multicolumn{3}{c|}{\textbf{SVM}} \\
\cline{3-11}
& & \textbf{FS Time} & \textbf{CV Time} & \textbf{\#Feat.} & \textbf{FS Time} & \textbf{CV Time} & \textbf{\#Feat.} & \textbf{FS Time} & \textbf{CV Time} & \textbf{\#Feat.} \\
\hline
\multirow{7}{*}{PCC} & GSA & 38.00 & 0.295 & 409.6 & 38.26 & 0.620 & 405.8 & 38.19 & 0.296 & 431.4 \\
 & BBA & 27.39 & 0.178 & 302.8 & 27.59 & 0.495 & 324.0 & 27.24 & 0.289 & 283.8 \\
 & GA & \textbf{4.86} & 0.118 & 298.6 & \textbf{4.85} & 0.498 & 298.6 & \textbf{4.86} & 0.240 & 298.6 \\
 & CS & 25.44 & \textbf{0.058} & \textbf{160.0} & 25.79 & 0.493 & 326.0 & 25.12 & \textbf{0.089} & \textbf{94.4} \\
 & GWO & 21.21 & 0.117 & 352.0 & 21.13 & \textbf{0.471} & 318.6 & 21.29 & 0.246 & 304.8 \\
 & PSO & 17.89 & 0.112 & 351.0 & 17.91 & 0.480 & 328.4 & 17.82 & 0.257 & 344.8 \\
 & WOA & 16.16 & 0.072 & 215.4 & 16.09 & 0.394 & \textbf{226.2} & 16.16 & 0.193 & 232.6 \\
\hline
\multirow{7}{*}{SCC} & GSA & 38.84 & 0.113 & 407.8 & 38.56 & 0.533 & 404.4 & 38.73 & 0.366 & 410.6 \\
 & BBA & 27.27 & 0.099 & 338.0 & 27.40 & 0.587 & 319.2 & 27.38 & 0.284 & 302.6 \\
 & GA & \textbf{4.87} & 0.122 & 298.6 & \textbf{4.87} & 0.498 & 298.6 & \textbf{4.87} & 0.234 & 298.6 \\
 & CS & 25.74 & 0.074 & 320.2 & 26.04 & 0.494 & 297.6 & 25.38 & \textbf{0.088} & 307.0 \\
 & GWO & 21.29 & 0.092 & 317.6 & 21.39 & \textbf{0.471} & 304.0 & 21.34 & 0.246 & 304.4 \\
 & PSO & 17.96 & 0.112 & 329.4 & 18.00 & 0.482 & 339.8 & 17.96 & 0.256 & 340.6 \\
 & WOA & 16.24 & \textbf{0.072} & \textbf{223.0} & 16.21 & 0.393 & \textbf{195.8} & 16.19 & 0.194 & \textbf{237.4} \\
\hline
\multirow{7}{*}{Relief} & GSA & 38.69 & 0.092 & 394.2 & 38.56 & 0.540 & 406.8 & 38.71 & 0.369 & 413.2 \\
 & BBA & 27.34 & 0.100 & 278.4 & 27.36 & 0.587 & 299.0 & 27.47 & 0.283 & 310.2 \\
 & GA & \textbf{4.90} & 0.122 & 298.6 & \textbf{4.89} & 0.497 & 298.6 & \textbf{4.90} & 0.234 & 298.6 \\
 & CS & 25.69 & \textbf{0.054} & \textbf{160.0} & 26.03 & 0.496 & \textbf{34.5} & 25.43 & \textbf{0.087} & \textbf{94.4} \\
 & GWO & 21.35 & 0.112 & 352.0 & 21.01 & 0.469 & 318.6 & 21.18 & 0.244 & 304.8 \\
 & PSO & 17.97 & 0.112 & 351.0 & 17.96 & 0.483 & 328.4 & 17.91 & 0.259 & 344.8 \\
 & WOA & 16.16 & 0.072 & 215.4 & 16.12 & \textbf{0.399} & 226.2 & 16.22 & 0.192 & 232.6 \\
\hline
\end{tabular}}
\end{table}

\subsection*{Statistical Test Results}
\begin{figure}[H]
    \centering
    \includegraphics[width=\textwidth]{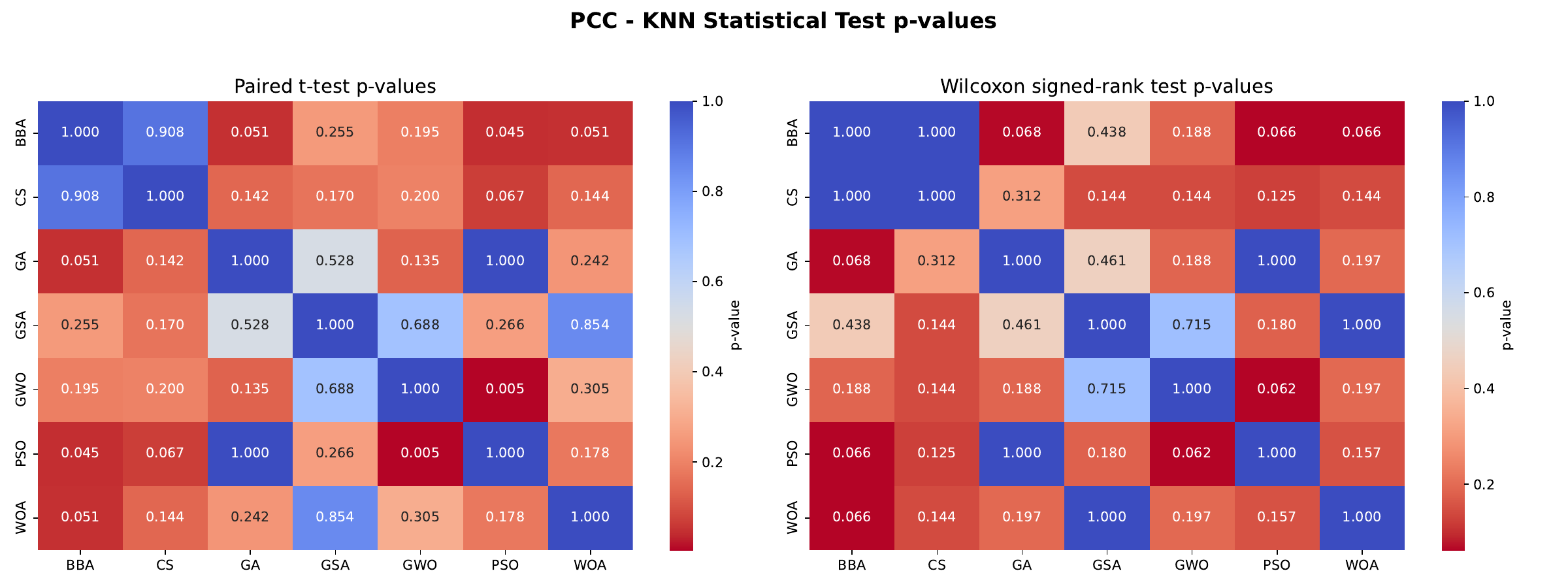}
    \includegraphics[width=\textwidth]{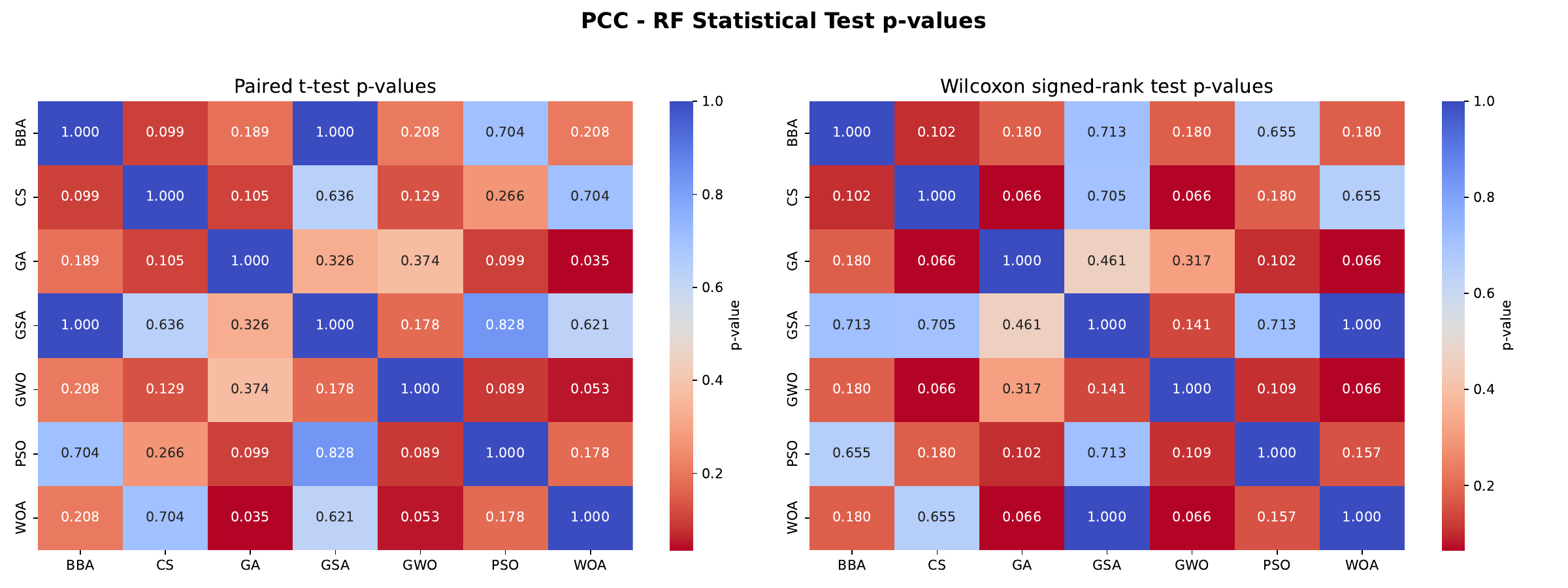}
    \includegraphics[width=\textwidth]{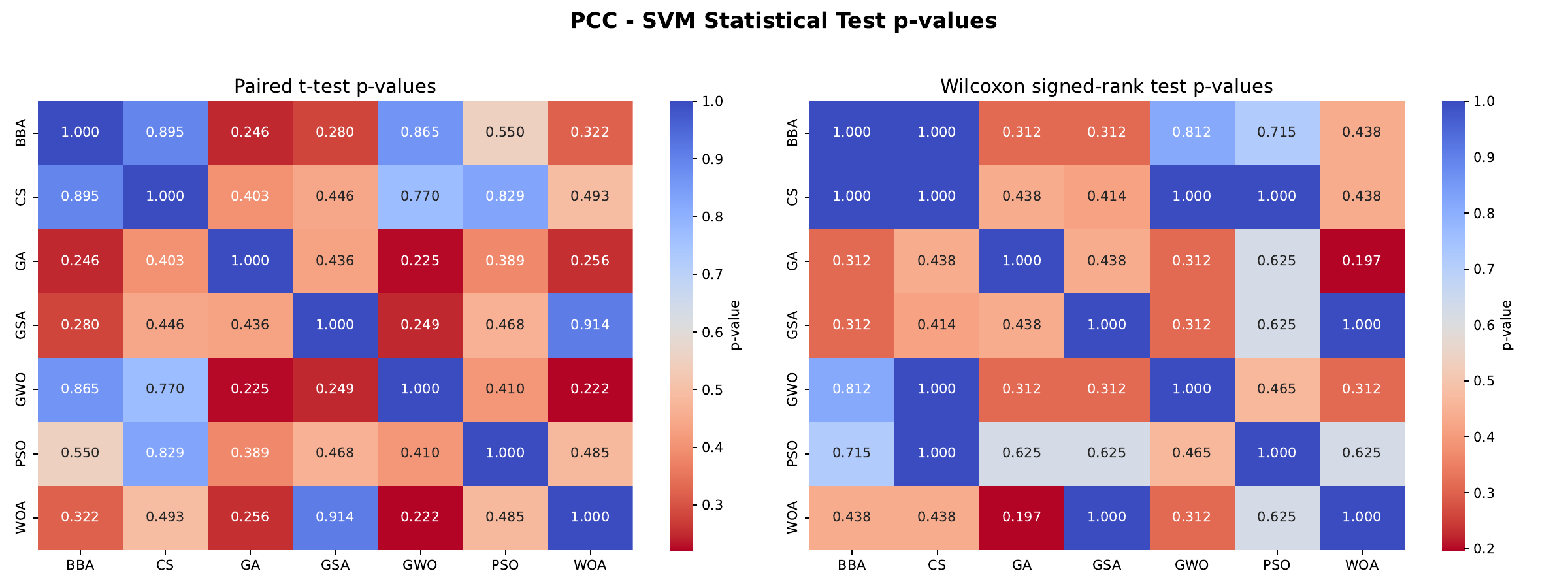}
    \caption{Statistical tests for PCC}
    \label{appendix:PCC}
\end{figure}

\begin{figure}[H]
    \centering
    \includegraphics[width=\textwidth]{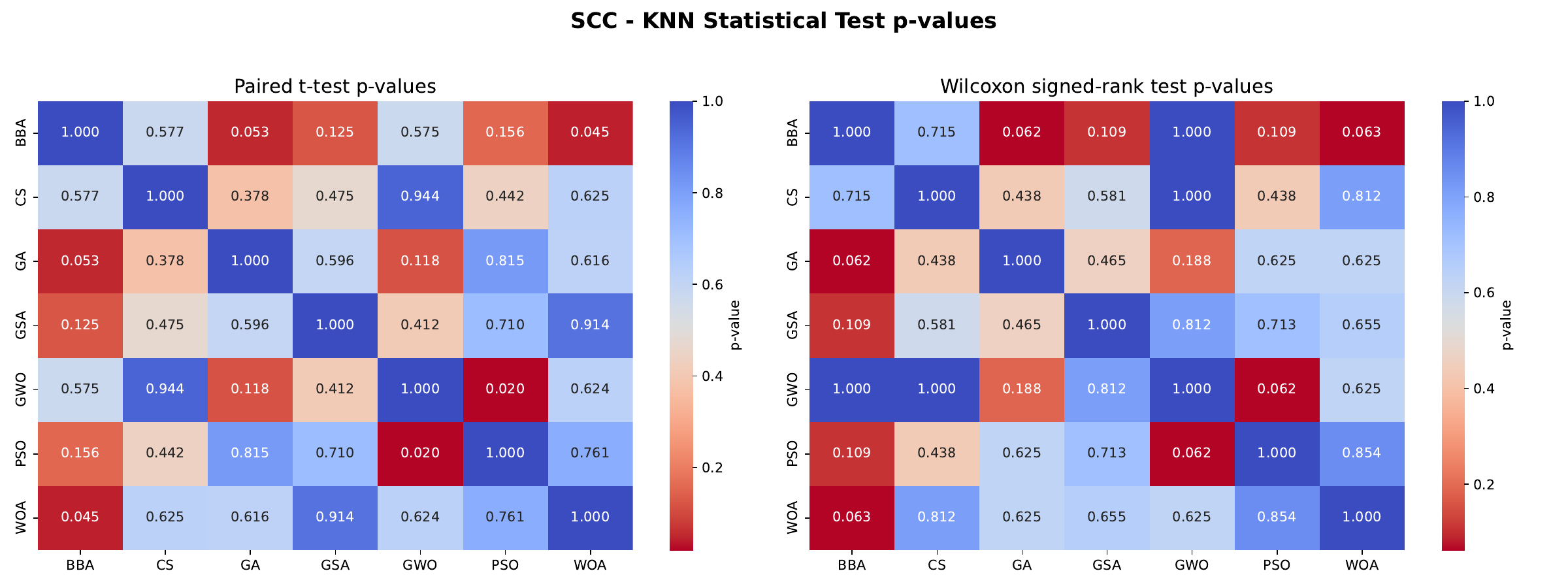}
    \includegraphics[width=\textwidth]{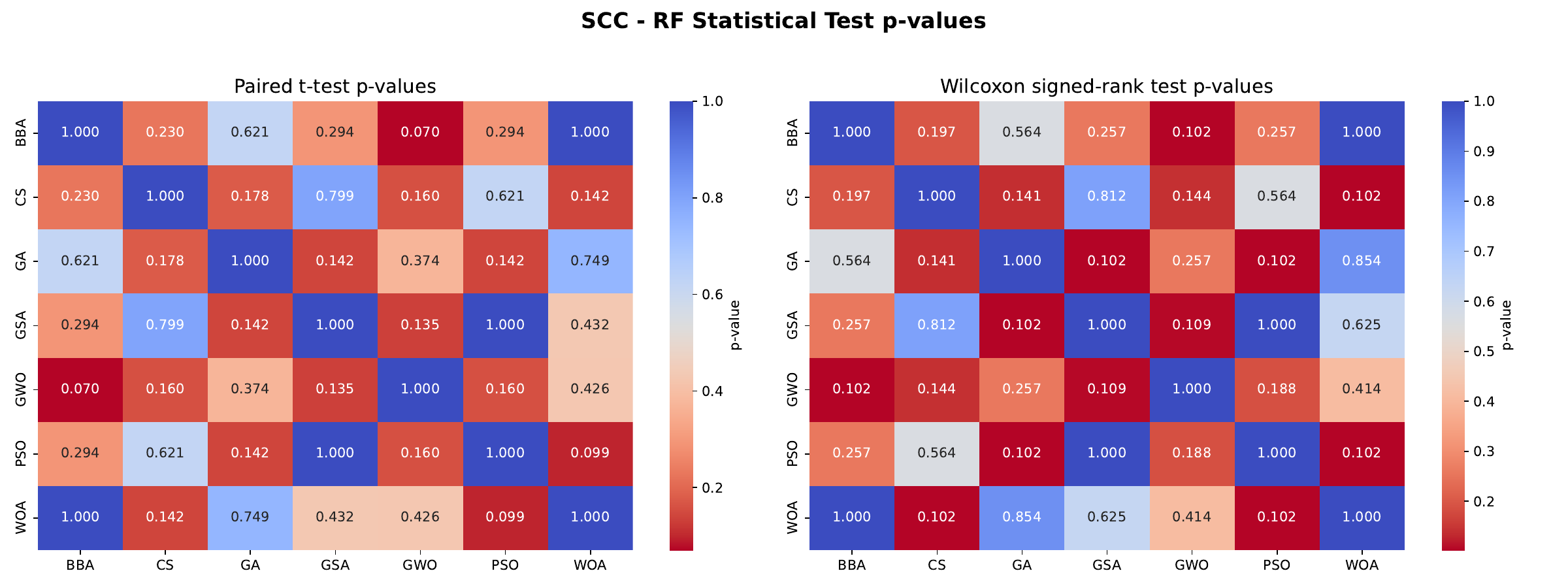}
    \includegraphics[width=\textwidth]{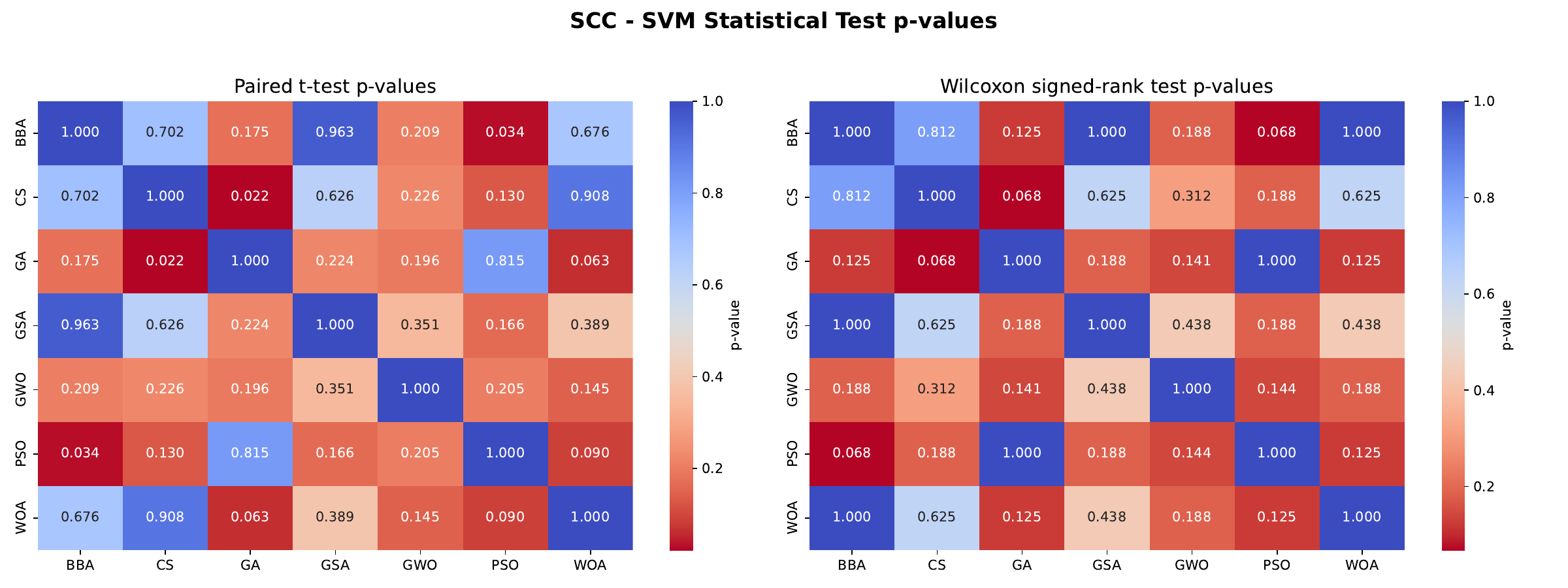}
    \caption{Statistical tests for SCC}
    \label{appendix:SCC}
\end{figure}

\begin{figure}[H]
    \centering
    \includegraphics[width=\textwidth]{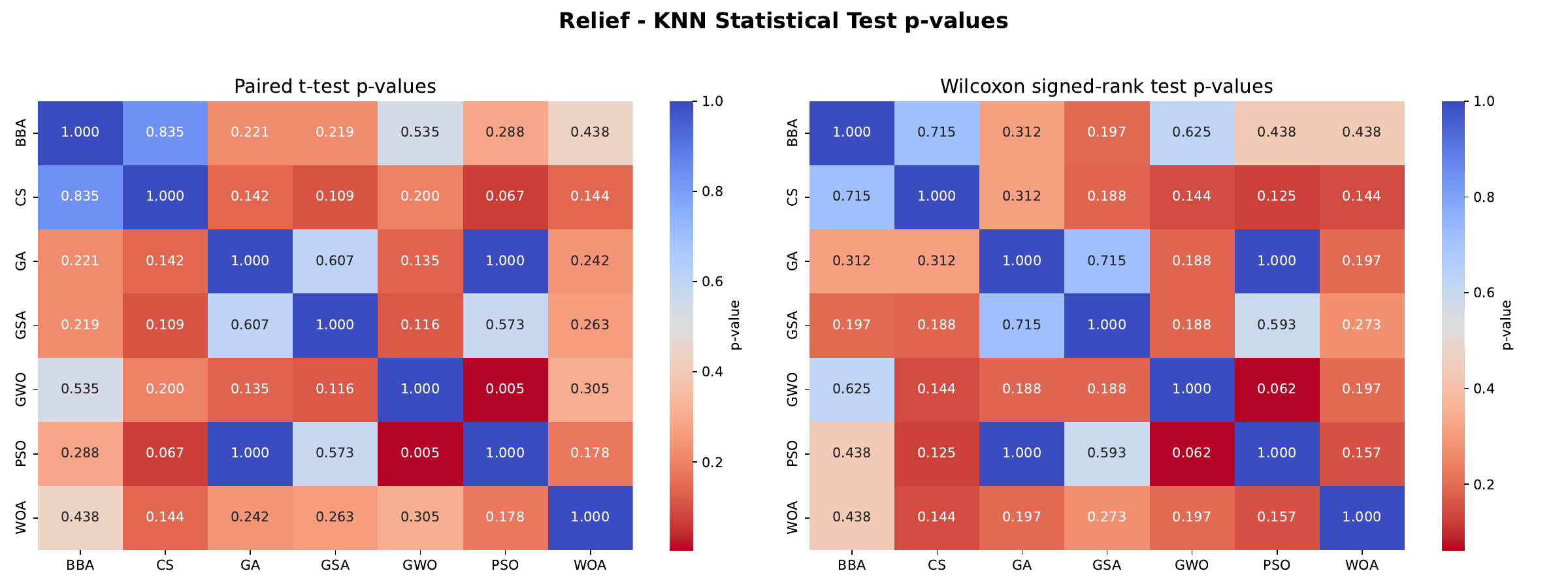}
    \includegraphics[width=\textwidth]{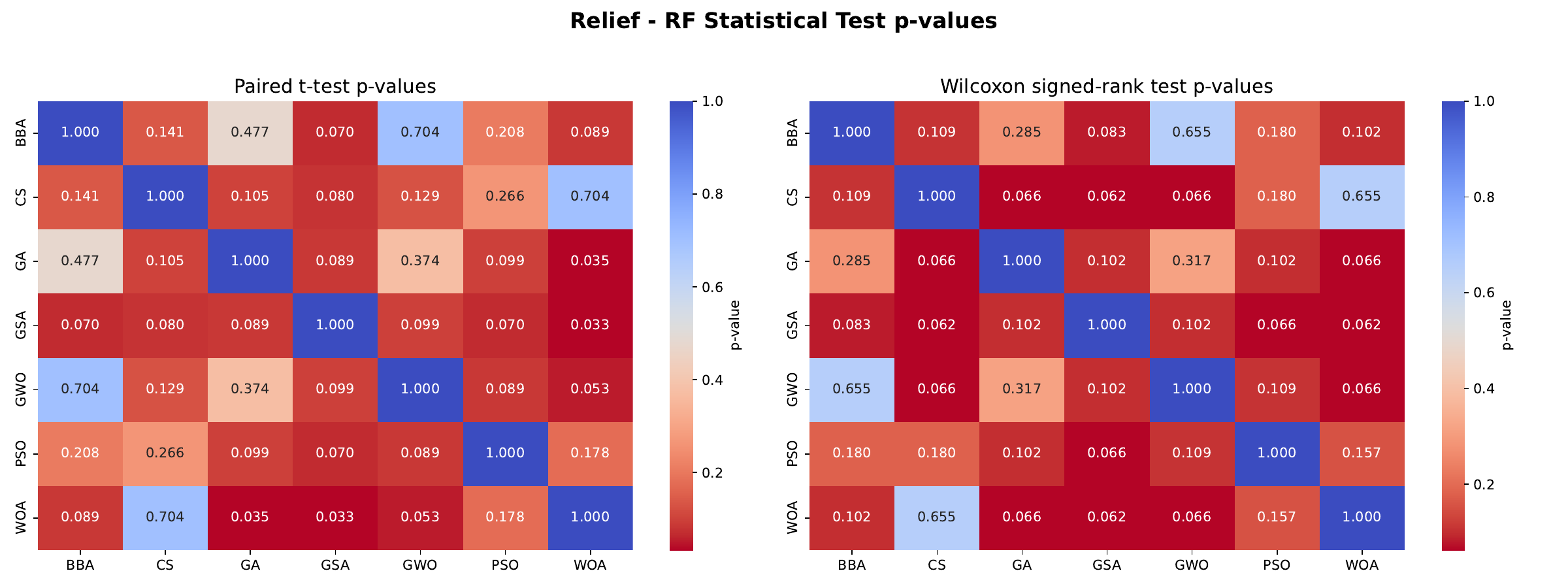}
    \includegraphics[width=\textwidth]{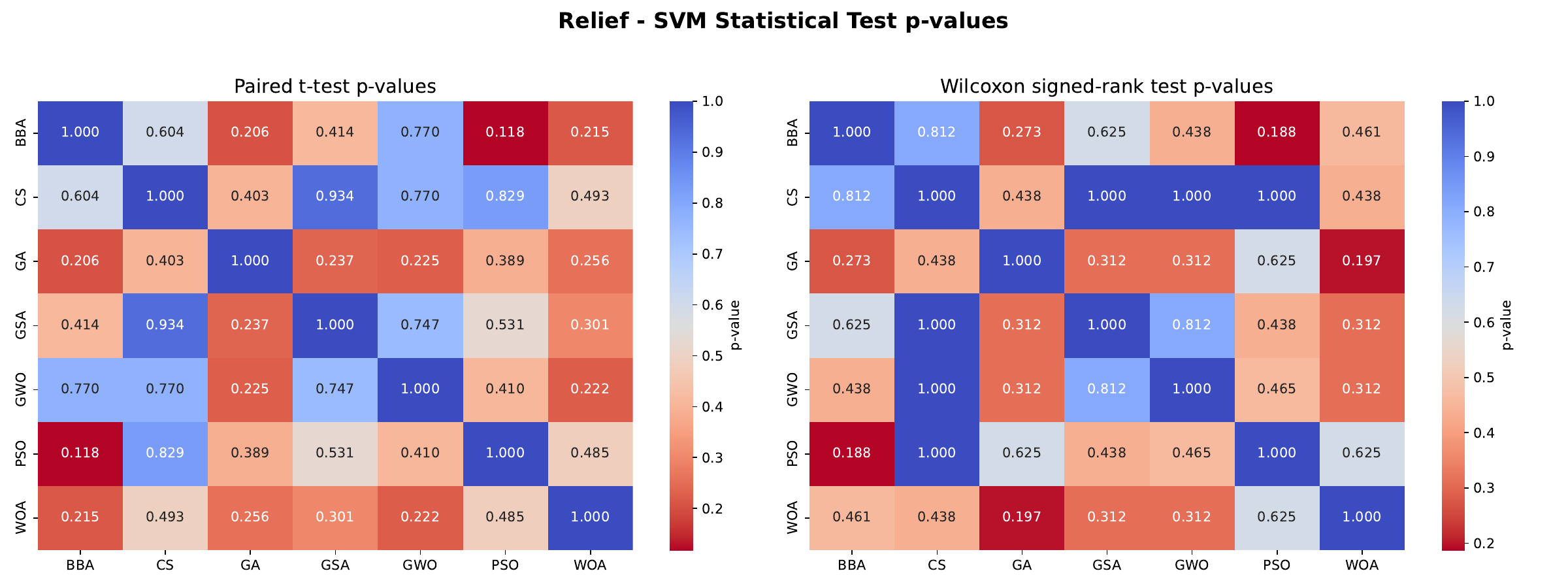}
    \caption{Statistical tests for Relief ranking coefficient}
    \label{appendix:Relief}
\end{figure}

From Figs.~\ref{appendix:PCC},~\ref{appendix:SCC}, and~\ref{appendix:Relief}, it can be inferred that most nature-inspired feature selection algorithms yield statistically similar classification performances when combined with KNN and SVM classifiers. However, certain pairings, particularly those involving the RF classifier, exhibit significant or borderline differences. Additionally, variations among the different feature selection algorithms are apparent. Across all cases, it is evident that the Relief ranking coefficients combined with the RF classifier consistently achieve the best performance.

\end{document}